\documentclass[twoside]{article}

\usepackage{tikz}
\usepackage{algorithm}
\usepackage[noend]{algpseudocode}
\usepackage{amsthm}
\usepackage{mathrsfs}
\newtheorem{definition}{Definition}

\usepackage[round]{natbib}

\usepackage{graphicx}
\usepackage{amsmath}
\usepackage{amssymb}
\usepackage{placeins}

\usepackage[preprint]{aistats2026}

\usepackage[utf8]{inputenc} 
\usepackage[T1]{fontenc}    
\usepackage{hyperref}       
\usepackage{url}            
\usepackage{booktabs}       
\usepackage{amsfonts}       
\usepackage{nicefrac}       
\usepackage{microtype}      
\usepackage{xcolor}         
\usepackage{stfloats}        
\usepackage{float}
\usepackage{cuted}

\makeatletter
\newif\ifpreprintclass
\@ifpackagewith{aistats2026}{preprint}
{
  \preprintclasstrue
}{
  \preprintclassfalse
}
\makeatother

\begin{document}

\twocolumn[

\aistatstitle{AlphaFold's Bayesian Roots in Probability Kinematics}
\aistatsauthor{%
  Thomas Hamelryck
  \And
  Kanti V. Mardia  
}
\aistatsaddress{
  University of Copenhagen, Denmark \\
  \And
  University of Leeds, UK
} 
]

\begin{abstract}
The seminal breakthrough of AlphaFold in protein structure prediction relied on a learned potential energy function parameterized by deep models, in contrast to its successors AlphaFold2 and AlphaFold3, which lack an explicit probabilistic interpretation. While AlphaFold’s potential was originally justified by heuristic analogy to physical {\em potentials of mean force}, we show that it can instead be understood as a principled instance of {\em probability kinematics} (PK), also known as {\em Jeffrey conditioning}, a generalization of Bayesian updating. This reinterpretation reveals that AlphaFold is a generalized Bayesian model that explicitly defines a posterior distribution over structures, providing a deeper explanation of its success and a foundation for future model design. To demonstrate this framework with precision, we introduce a tractable synthetic model in which an angular random walk prior is updated with distance-based evidence via PK, directly mirroring AlphaFold’s mechanism. This setting allows us to explore the probabilistic foundations of AlphaFold in a clear and interpretable way. Our work connects a landmark in protein structure prediction to a broader class of compositional deep generative models and points to new opportunities for principled probabilistic approaches.
\end{abstract}

\section{INTRODUCTION}

The protein structure prediction problem -- predicting the three-dimensional
(3D) structure of a protein from its amino acid sequence -- has
been a central unsolved challenge for more than
50 years \citep{dill2012protein}. In 2024, the shared Nobel Prize
in Chemistry was awarded to John Jumper and Demis Hassabis for AlphaFold's solution of this problem \citep{callaway2024chemistry}. Here, we focus on the initial formulation of AlphaFold, which we will refer to as AlphaFold1 (AF1) \citep{AF12019protein,af1_nature_2020}. AF1 makes use of a potential function parameterized by deep learning and (as we show here) rooted in generalized Bayesian updating, which differs considerably from the end-to-end deep learning pipeline approach of its two successors that despite increased pragmatic accuracy lack an explicit, principled, SE(3)-invariant probabilistic model over the conformational space of proteins \citep{jumper2021highly,abramson2024accurate, jing2025ai}. Predictions by AF1 are obtained by minimizing this potential energy function with respect to the protein's dihedral angles. We argue that AF1's probabilistic nature is well worth reconsidering as a basis for further development, as we discuss below.  

AF1's potential function was justified heuristically by appealing to physical \emph{potentials
of mean force} (PMFs) \citep{AF12019protein,af1_nature_2020} as used
in the physics of liquids, first formulated by Kirkwood in 1935 \citep{kirkwood1935}.
Here, we instead interpret AF1 as an instance of an under-recognized
generalization of Bayesian updating called \emph{Jeffrey conditioning}
or \emph{probability kinematics} (PK). PK was introduced by the philosopher
of logic and probability Richard C. Jeffrey in 1957 \citep{jeffrey1957contributions,jeffrey1965logic,jeffrey2004subjective},
and specifies how to incorporate {\em soft evidence}, i.e., 
updated probabilities on a partition of a random variable's sample space (see Figure
\ref{fig:PK-explained}), rather than conditioning on a concrete event. AF1 updates
a prior distribution over a protein's dihedral angles with soft evidence
on pairwise distances between amino acids, both obtained from deep
models. 

\begin{figure*}[t]
\centering
\begin{minipage}{0.33\linewidth}
    \centering

    \begin{tikzpicture}
    \draw[thick] (0,0) rectangle (6,4);

    \draw[thick, dashed] (3,2) ellipse (2.5 and 1.5); 
    \draw[thick, dashed] (3,2) ellipse (1.5 and 0.9); 
    \draw[thick, dashed] (3,2) ellipse (0.8 and 0.5); 

    \fill[color=gray, opacity=0.2] (0,0) -- (1.5,1) -- (2,2) -- (0,2) -- cycle;
    \draw[thick] (0,0) -- (1.5,1) -- (2,2) -- (0,2) -- cycle;
    \node at (0.6,1) {$\xi_1$}; 
    
    \fill[color=yellow, opacity=0.2] (1.5,1) -- (6,0) -- (6,1.5) -- (4,2) -- (2,2) -- cycle;
    \draw[thick] (1.5,1) -- (6,0) -- (6,1.5) -- (4,2) -- (2,2) -- cycle;
    \node at (5.5,1) {$\xi_2$}; 
    
    \fill[color=green, opacity=0.2] (4,2) -- (6,1.5) -- (6,4) -- (4.5,3.5) -- cycle;
    \draw[thick] (4,2) -- (6,1.5) -- (6,4) -- (4.5,3.5) -- cycle;
    \node at (5.5,3) {$\xi_3$}; 
    
    \fill[color=red, opacity=0.2] (0,2) -- (2,2) -- (2.5,3) -- (1,3.5) -- (0,4) -- cycle;
    \draw[thick] (0,2) -- (2,2) -- (2.5,3) -- (1,3.5) -- (0,4) -- cycle;
    \node at (0.4,3.0) {$\xi_4$}; 
    
    \fill[color=blue, opacity=0.2] (2,2) -- (4,2) -- (4.5,3.5) -- (2.5,3) -- cycle;
    \draw[thick] (2,2) -- (4,2) -- (4.5,3.5) -- (2.5,3) -- cycle;
    \node at (4,3.0) {$\xi_5$}; 

    \node at (2.5,3.7) {$\xi_6$}; 
    \node at (2.5,0.25) {$\xi_6$}; 

    \end{tikzpicture}   
\end{minipage}
\begin{minipage}{0.5\linewidth}
    \centering
    \begin{flushright}
    \begin{align*}
    (A)\,\,\,\, & \pi(\boldsymbol{\omega})=\sum_{i}\pi(\boldsymbol{\omega}\mid\xi_{i})\pi(\xi_{i}),\\
    (B)\,\,\,\, & p(\boldsymbol{\omega}\mid\xi_{i})=\pi(\boldsymbol{\omega}\mid\xi_{i})\\
    (C)\,\,\,\, & \Rightarrow p(\boldsymbol{\omega})=\sum_{i}\pi(\boldsymbol{\omega}\mid\xi_{i})p(\xi_{i})\\
    (D)\,\,\,\, & \Rightarrow p(\boldsymbol{\omega})=\frac{p\left(\xi(\boldsymbol{\omega})\right)}{\pi\left(\xi(\boldsymbol{\omega})\right)}\pi(\boldsymbol{\omega})
    \end{align*}
    \par\end{flushright}
\end{minipage}
\caption{\textbf{Schematic example of generalized Bayesian updating by PK.} The dotted lines denote the equiprobability
contours of the probability density function (PDF) of a prior distribution, $\pi(\boldsymbol{\omega})$, in this example
defined on a rectangle. The solid lines represent a finite partition $\{\xi_{i}\}$
of the sample space of $\pi(\boldsymbol{\omega})$, with probabilities
$\{\pi(\xi_{i})\}$. $\pi(\boldsymbol{\omega})$ can thus be written as in (A). New evidence is then obtained in the form of updated
probabilities for the partition,$\left\{ p(\xi_{i})\right\} $. PK updates the prior $\pi(\boldsymbol{\omega})$ to the posterior $p(\boldsymbol{\omega})$ by keeping the conditional following (B), which is called the {\em Jeffrey-condition} or {\em J-condition}, and replacing the marginal as in (C). An equivalent expression (D), where $\xi(\boldsymbol{\omega})$ is the partition element that contains $\boldsymbol{\omega}$, is obtained by applying Bayes' formula to (C). The latter form, called the {\em reference ratio} update, is of interest when
the conditional in (C) is not readily available or tractable, as is the case for AF1. PK readily generalizes to infinite partitions (see Section \ref{sec:Probabilistic-framework}).}
\label{fig:PK-explained}
\end{figure*}

This theoretical paper offers the following contributions:
\begin{itemize}
\item Building on a definition of PK for probability density functions (PDFs)
and infinite partitions tailored to the case of AF1, we propose a
generalized Bayesian reinterpretation of AF1´s potential
function, and make explicit its functional form as a prior density adjusted by a ratio of PDFs. 
\item We contrast our framework with AF1´s heuristic explanation
based on PMFs and link its deep model to other statistical models of protein
structure. This bridge is timely, as there is a growing need for
modeling epistemic and aleatoric uncertainty \citep{bowman2024alphafold,ramasamy2025assessing,chakravarty2024alphafold,agarwal2024power,gavalda2025gradations,chakravarty2025proteins,lopez2025severe}.
\item We introduce a tractable two-dimensional synthetic model that is analogous
to AF1's potential. The model updates an angular prior with evidence in the form of a PDF over a distance. This setup mirrors the probabilistic framework
behind AF1's potential and allows us to demonstrate
the update's precision via hypothesis testing. In the Appendix, we also provide an application to a small 3D protein model. 
\item We reveal PK as a previously unrecognized foundation for formulating deep generative models of biomolecular structure.
\item Our results show PK's relevance and potential in 
formulating {\em compositional} deep generative models, i.e., models constructed by combining simpler probabilistic components into a coherent whole \citep{ghahramani2015probabilistic,wood2014new,bingham2019pyro}.
\end{itemize}

Our aim is not to improve or assess empirical accuracy, as AF1’s historical success is well established \citep{AF12019protein, af1_nature_2020}, but to clarify its previously unrecognized statistical foundation and point out its potential for future work. To do this, we make use of a principled probabilistic framework, tested in a controlled synthetic setting where accuracy can be assessed.   

\FloatBarrier 

\section{BACKGROUND\label{sec:Background}}

\subsection{Preliminaries\label{sec:Assumptions-and-Notation}}

\begin{figure}[t]
\begin{centering}
\includegraphics[width=1\columnwidth]{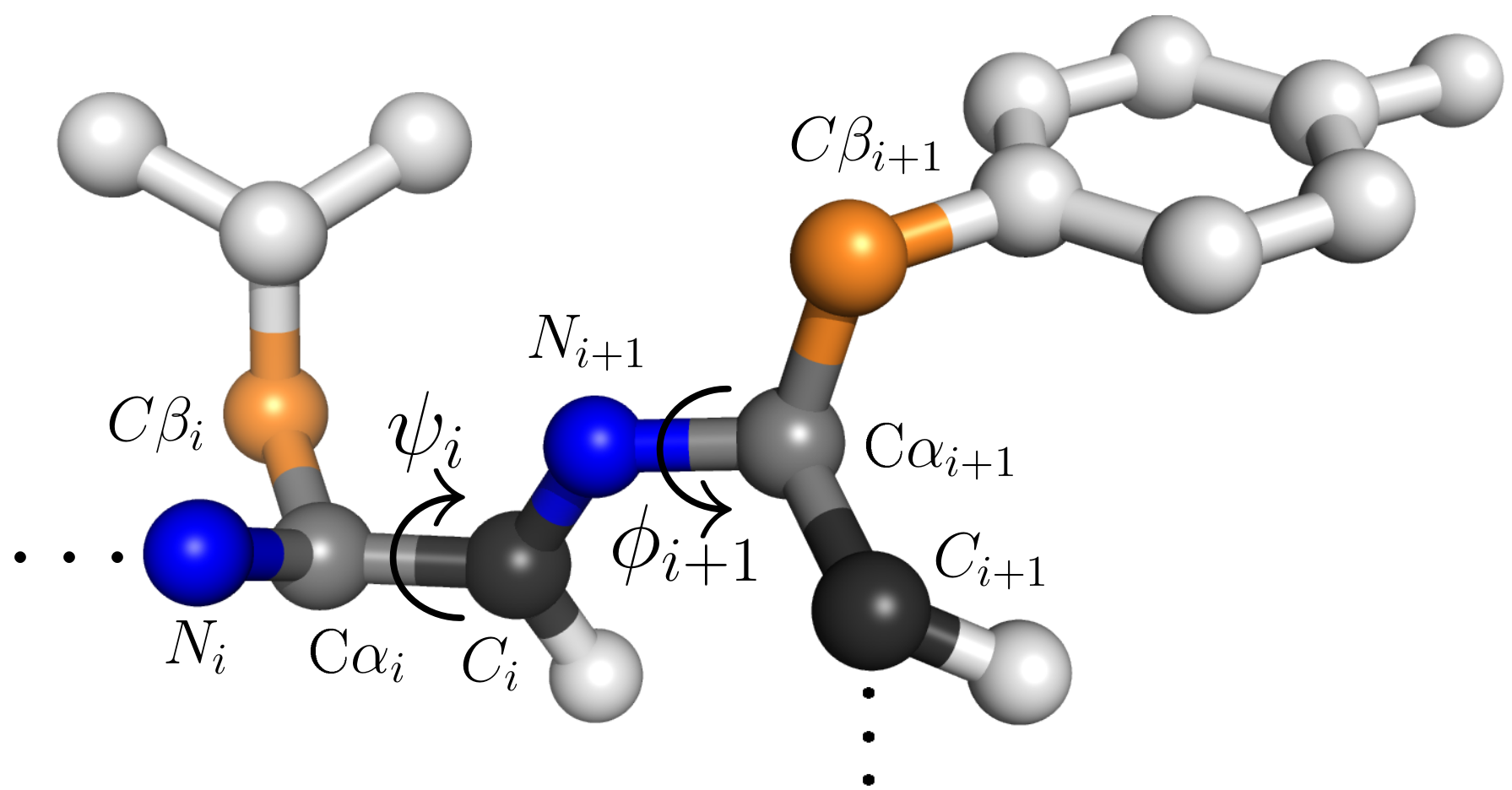}
\hfill
\includegraphics[width=0.6\columnwidth]{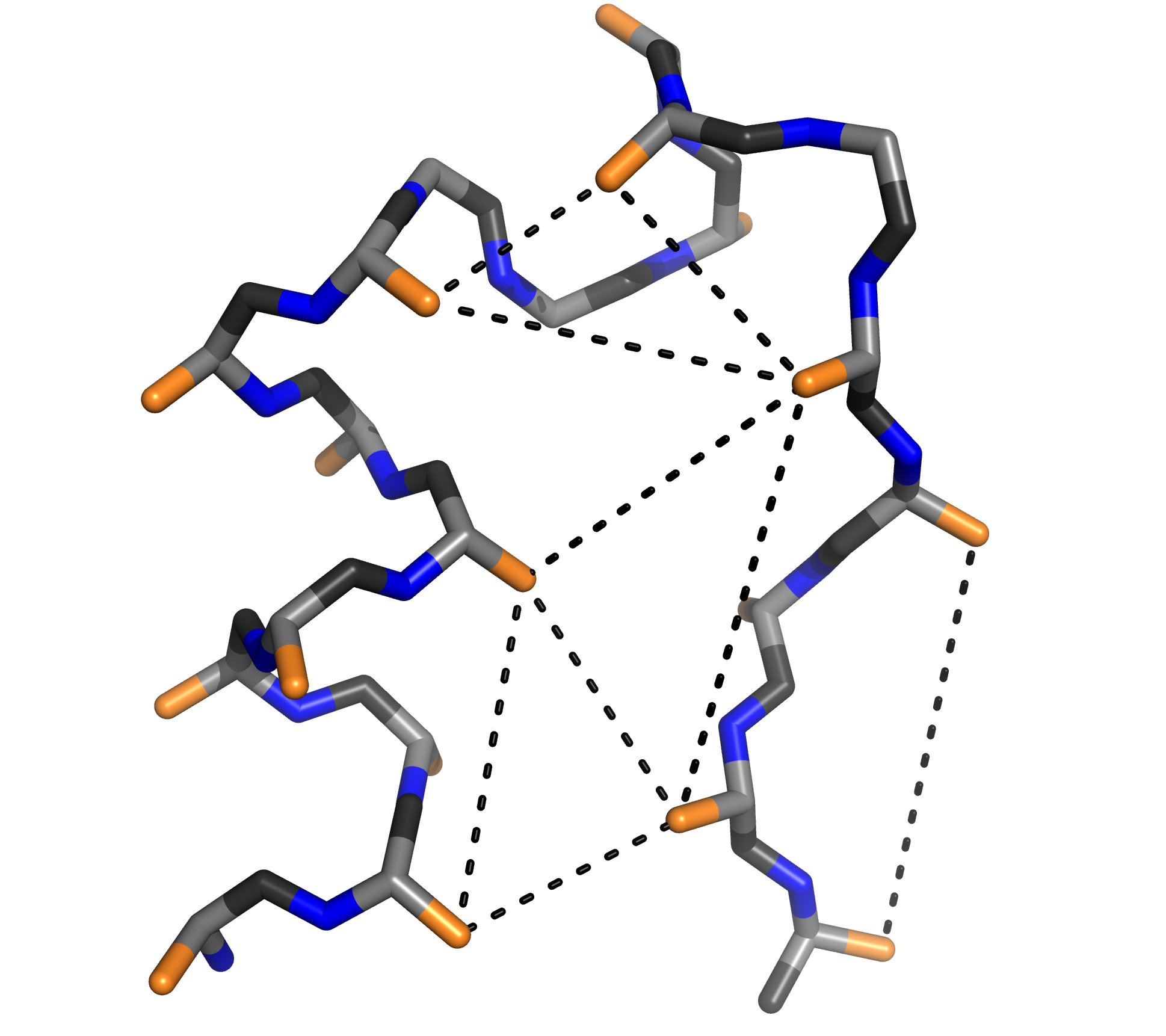}
\par\end{centering}
\caption{\textbf{(top)} Two amino acids in a protein, with relevant atoms
and dihedral angles labeled. Atoms that do not play a role in the
AF1 potential, including side chain and backbone oxygen atoms, are
shown in white. The continuation of the protein backbone is indicated
with dots.\textbf{ (bottom}) A protein 
with some $C\beta-C\beta$ distances shown as dotted lines.
Only the atoms relevant to the AF1 potential are shown. Figures made with PyMol \citep{PYMOL}.
\label{fig:protein}}
\end{figure}

The protein amino acid sequence, $\boldsymbol{s}$, of length $L,$
is represented by a $L \times 20$ tensor of one-hot encodings of the 20 amino acid symbols. $\boldsymbol{M}$ is an
$M\times L \times 20$ tensor that contains the multiple sequence alignment
of $M$ protein sequences related to $\boldsymbol{s}$. Following AF1, a protein
structure will be represented by its backbone atoms only, including
$N,C\alpha,C,C\beta$ atoms (see Figure \ref{fig:protein}). The 3D coordinates
of these four atoms for each amino acid are in the tensor
$\boldsymbol{X}$, which is $L\times4\times3$. 

Protein structure is often parameterized
using dihedral angles \citep{ramachandran1963stereochemistry},\footnote{A {\em dihedral angle} or {\em torsion angle}, defined by four consecutively bonded atoms, is the angle between the planes formed by the first three atoms and the last three. Under the common assumption that bond lengths and bond angles take fixed, ideal values, a protein's 3D structure is fully determined by its dihedral angles.} which has the advantage of invariance to rotation and translation and statistical simplicity \citep{edgoose1998mml, Torus08, hamelryck2012bayesian}.
Let $\boldsymbol{\phi}$ and $\boldsymbol{\psi}$ be vectors of
$L$ dihedral angles in $(-\pi ,\pi]$. We will write $\boldsymbol{X}(\boldsymbol{\phi},\boldsymbol{\psi})$
for the realization of $\boldsymbol{X}$ up to an arbitrary
rotation and translation for a given set of dihedral angles. Following AF1 \citep{AF12019protein,af1_nature_2020}, we assume
ideal bond angles and lengths, {\em trans} peptide bonds, and ignore amino acid side chains and backbone
oxygen atoms, without any loss of generality (see Figure \ref{fig:protein}). 

Let $\boldsymbol{D}$ represent the $L\times L$ distance matrix of distances
between $L$ $C\beta$ atoms, with $L(L-1)/2$ unique values. Note that
although these pairwise distances only involve the $C\beta$ atoms,
they are taken as representative distances between amino acids. 
A given set of atomic coordinates, $\boldsymbol{X}$, or dihedral angles, $(\boldsymbol{\phi},\boldsymbol{\psi})$, determines $\boldsymbol{D}$ fully
-- we will write $\boldsymbol{D}(\boldsymbol{X})$ and $\boldsymbol{D}(\boldsymbol{\phi},\boldsymbol{\psi})$ for the 
realization of $\boldsymbol{D}$ for given coordinates or dihedral angles. 

We will use the following generic notation to emphasize features common to PK and PMFs. 
Let $\boldsymbol{\omega}$ be a random variable, and let 
$\boldsymbol{\xi}$ be a random variable obtained via a 
many-to-one mapping of $\boldsymbol{\omega}$. 
Thus, $\boldsymbol{\xi}$ is the random variable and $\boldsymbol{\xi}(\boldsymbol{\omega})$ is its concrete \emph{realization} 
for a given value of $\boldsymbol{\omega}$. We will call $\boldsymbol{\omega}$ a {\em fine-grained} random variable, and  $\boldsymbol{\xi}$ a corresponding {\em coarse-grained} or {\em collective} random variable. 
Depending on the context, $\boldsymbol{\omega}$ and $\boldsymbol{\xi}$ may be matrices, vectors, sets or scalars; 
in the latter two cases we omit the boldface. These symbols are reused and explained in their context as needed. Notably, in the previous paragraph, $\boldsymbol{D}$ is a coarse grained, collective variable, while $\boldsymbol{X}$ is a corresponding fine-grained variable, corresponding to instances of $\boldsymbol{\xi}$ and $\boldsymbol{\omega}$, respectively.
 Appendix (\ref{sec:appendix_notation}) gives an overview of notation and abbreviations.

\subsection{AF1's potential}

AF1's potential contains four terms, three of which are parameterized
by deep models. A fourth, repulsive potential is derived from physical considerations
and merely serves to prevent atoms from overlapping. The first three
potentials are central to this article. Section \ref{sec:AF1's-Potential-Re-evaluated} provides details on the PDFs mentioned below. 

The first potential concerns the dihedral angles and is given by
\begin{equation}
V_{\textrm{A}}(\boldsymbol{\phi},\boldsymbol{\psi})=-\sum_{i}\log p\left(\phi_{i},\psi_{i}\mid\boldsymbol{s},\boldsymbol{M}\right),\label{eq:AF1_angles}
\end{equation}
where the sum runs over all amino acids. The second energy term contains the pairwise distances between the
$C\beta$ atoms, which serve as representative distances between amino
acids $i$ and $j$,
\begin{equation}
V_{\textrm{D}}(\boldsymbol{D})=-\sum_{i,j,i\neq j}\log p\left(D_{i,j}\mid\boldsymbol{s},\boldsymbol{M}\right).\label{eq:AF1_distance}
\end{equation}
As we will show in Section \ref{sec:AF1's-Potential-Re-evaluated},
the formulation of a consistent joint probability distribution over
angles and distances requires a third term -- a so-called \emph{reference
potential}, which also concerns pairwise distances. In AF1, this potential
depends only on the number of amino acids in the protein, $L$,
and a one-hot vector that specifies the positions of Glycine residues,
$\mathbb{I}_{\textrm{Gly}}(\boldsymbol{s})$. Glycine is a small
amino acid that allows exceptional conformational freedom to the protein
chain. The reference potential is 
\begin{equation}
V_{\textrm{R}}(\boldsymbol{D})=-\sum_{i,j,i\neq j}\log p\left(D_{i,j}\mid L,\mathbb{I}_{\textrm{Gly}}(\boldsymbol{s})\right).\label{eq:AF1_reference}
\end{equation}
What AF1 refers to as a Kirkwood PMF is then obtained from their
difference, 
\begin{equation}
 V_{\textrm{PMF}}\left(\boldsymbol{D}\right)\text{=}V_{\textrm{D}}\left(\boldsymbol{D}\right)-V_{\textrm{R}}\left(\boldsymbol{D}\right).\label{eq:AF1_ratio}
\end{equation}
The final potential brings the three potentials above together, and
combines it with the predefined repulsive potential, 
\begin{multline}
V(\boldsymbol{\phi},\boldsymbol{\psi})=V_{\textrm{PMF}}\left(\boldsymbol{D}(\boldsymbol{\phi},\boldsymbol{\psi})\right)+V_{\textrm{A}}(\boldsymbol{\phi},\boldsymbol{\psi}) \\ +V_{\textrm{Repulsive}}\left(\boldsymbol{X}(\boldsymbol{\phi},\boldsymbol{\psi})\right).\label{eq:AF1_final_potential}
\end{multline}

Note that $\boldsymbol{D}$ and $\boldsymbol{X}$ (up to an arbitrary rotation and translation) in the above expression are fully determined by the dihedral angles, $\boldsymbol{\phi}$ and $\boldsymbol{\psi}$ (see Figure \ref{fig:protein}). Prediction by AF1 amounts
to minimizing the potential as a function of the dihedral angles using
gradient descent. In the next section, we explain how AF1 justifies
this potential by heuristic analogy with PMFs for liquid systems. 

\subsection{AF1's potential justified as a PMF}

\paragraph{Physical potentials of mean force}

A PMF can be understood as an \emph{averaged},
\emph{mean} or \emph{effective} potential regarding
some coordinates (typically called the \emph{collective} or \emph{coarse-grained variables}) of a physical
system, while averaging over the remaining degrees of freedom \citep{chandler1987,zuckerman2010statistical,hartmann_pmf2011}.
A PMF is defined as 
\begin{align}
p(\boldsymbol{\xi}) & \propto\exp(-\textrm{PMF}(\boldsymbol{\xi})/kT) \\
                    & =\int\delta\left(\boldsymbol{\xi}-\boldsymbol{\xi}(\boldsymbol{X})\right)\exp(-U(\boldsymbol{X})/kT)\textrm{d}\boldsymbol{X},\label{eq:PMF_delta-1}
\end{align}
where $k$ is Boltzmann's constant, $T$ is the temperature, $\boldsymbol{X}$
are the atomic coordinates of the system, $\boldsymbol{\xi}$ is the collective random
variable (e.g., distances between a subset of the atoms), $\boldsymbol{\xi}(\boldsymbol{X})$ is the realization of $\boldsymbol{\xi}$ at $\boldsymbol{X}$, $\delta(.)$ is the Dirac delta function, and the exponential
factor is the Boltzmann distribution associated with the internal energy
$U$. 

PMFs were introduced by Kirkwood in his seminal paper from 1935 \citep{kirkwood1935},
applied to mixtures of liquids. For this particular physical system, Kirkwood proposed expressions
of the form,
\begin{equation}
W_\textrm{Kirkwood}(r)=-kT\log\left(\frac{p(r)}{p_{\textrm{Reference}}(r)}\right),\label{eq:Kirkwood}
\end{equation}
where $r$ is the distance between two identified liquid particles. The numerator in the log
is the probability density observed in a liquid, while the denominator
refers to a so-called {\em reference distribution}. In the case of Kirkwood
potentials, the reference distribution is well-defined, corresponding to
an ideal state of non-interacting particles \citep{kirkwood1935}. It serves to isolate genuine intermolecular 
interactions from features that are purely geometric or entropic. Technically, a Kirkwood potential is thus a \emph{relative} PMF, corresponding to a difference between the PMFs of the interacting liquid and its corresponding non-interacting reference system. We now turn to the heuristic
adoption of formulas resembling Kirkwood potentials, as done by AF1, for biomolecular prediction
and simulation. 

\paragraph{Knowledge based potentials}

Knowledge based potentials (KBPs) attempt to formulate empirical energy
functions based on statistics derived from the database of known structures
\citep{pohl1971empirical,tanaka1976medium,miyazawa1985estimation,sippl_pmf1990}. KBPs
based on the pairwise distances between $C\beta$ atoms were introduced
by \cite{sippl_pmf1990} and widely adopted \citep{sippl1995knowledge,sippl1993boltzmann,sippl1996helmholtz}. The
energy of a protein structure following such a KBP is given by 
\begin{equation}
\Delta E_{\boldsymbol{s}}(\boldsymbol{X})=\sum_{i<j}-\log\left(\frac{p\left(D_{i,j}(\boldsymbol{X})\mid s_{i},s_{j}\right)}{p_{\textrm{R}}\left(D_{i,j}(\boldsymbol{X})\mid s_{i},s_{j}\right)}\right),\label{eq:Sippl}
\end{equation}
where $s_i, s_j$ are amino acid labels. 
These heuristic potentials were loosely inspired by the ratio of PDFs arising in Kirkwood potentials (Eq. \ref{eq:Kirkwood}).

Early Rosetta work \citep{Rosetta1997} used a Sippl-style potential involving discretized distances combined with Monte Carlo sampling of protein fragments. Sippl-style potentials were also used for protein structure evaluation, fold recognition and threading, and model refinement \citep{zhou2002distance, shen2006statistical, lu2008opus, zhang2010novel}. In the 2010s, later versions of Rosetta and other programs instead typically relied on predicted contacts/distances as straightforward restraints, which is distinct from Sippl-style potentials and probability kinematics.

Despite pragmatic success, their
theoretical justification as genuine PMFs was widely contested \citep{pmid8609636,ben-naim:3698,pmid9526121}. Notably,
unlike physical PMFs and Kirkwood potentials, KBPs are not derived from a Boltzmann distribution of a single physical system, but from a heterogeneous data set of different protein
structures, and the reference distribution is undefined \citep{ben-naim:3698,pmid9526121}. Attempts at \emph{qualitative, heuristic}
justification invoked properties of random polymers \citep{finkelstein1995protein}
and probabilistic reasoning \citep{Rosetta1997}. Subsequently,
KBPs were \emph{quantitatively, formally} explained \citep{Hamelryck_PMF2010,valentin2014},
and eventually recognized as an instance of PK \citep{hamelryck_PK_Book2015},
to which we turn next. 

\subsection{Probability kinematics\label{subsec:Probability-kinematics}}

As in the classic example of {\em Whitworth's horses} \citep{diaconis1982updating} (see Appendix \ref{sec:appendix_whitworth}), new information often does
not arrive in the form of a deterministic event, but rather as a revision
in the probability of various hypotheses. PK provides a framework
for updating beliefs under such uncertain evidence. Unlike classical
Bayesian conditioning, which updates a prior $\pi(\boldsymbol{\omega})$ to a posterior
$p(\boldsymbol{\omega})\propto p(\boldsymbol{y}\mid\boldsymbol{\omega})\pi(\boldsymbol{\omega})$
using data $\boldsymbol{y}$, PK updating uses revised probabilities
on a partition of the sample space of $\boldsymbol{\omega}$ (Figure \ref{fig:PK-explained}).

Given a prior distribution $\pi(\boldsymbol{\omega})$ over a sample
space $\Omega$, and a finite partition $\{\xi_{i}\}$ of $\Omega$ 
along with new marginal probabilities ${p(\xi_{i})}$ that differ from the prior probabilities ${\pi(\xi_{i})}$, the PK update prescribes
the posterior as
\begin{equation}
p(\boldsymbol{\omega})=\sum_{i}\pi(\boldsymbol{\omega}\mid \xi_{i})p(\xi_{i}),\label{eq:PK-finite-case}
\end{equation}
by construction making use of the \emph{Jeffrey- or J-condition} $p(\boldsymbol{\omega}\mid \xi_{i})=\pi(\boldsymbol{\omega}\mid \xi_{i})$.

Unlike conventional Bayesian inference, PK does not rely on an explicit
likelihood function, and its evidence is not a concrete event. Nonetheless, PK preserves essential features
of Bayesian updating. First, PK satisfies the Dutch book argument
\citep{armendt1980there}. Second, PK reduces to conventional Bayesian
updating when for one of the partition elements,  $p(\xi_i)=1$. Third, among all distributions that match the updated marginals
$\{p(\xi_{i})\}$, the PK update minimizes the Kullback-Leibler divergence
to the prior $\pi$ \citep{diaconis1982updating}. This links the method to a variational principle \citep{blei2017variational,kingma2019introduction}
that fits the Bayesian interpretation of probability \citep{diaconis1982updating}.
PK is usually formulated in terms of probabilities and finite partitions \citep{meehan2020jeffrey}.
In the next section, we outline the more general probabilistic framework
needed to address AF1.

\section{PROBABILISTIC FRAMEWORK\label{sec:Probabilistic-framework}}

In this section, we first formulate a definition of PK that is suitable
for our purposes (i.e., concerning PDFs and infinite partitions), and review some of PK's properties. As a preliminary
to its application in the case of AF1, we consider the case of pairwise
distance KBPs and contrast PK with physical PMFs. 

\subsection{PK for the case of PDFs and infinite partitions}

We provide a definition tailored to the present case, based
on \citep{diaconis1982updating} and \citep{meehan2020jeffrey}, while
avoiding the formalism of measure theory and irrelevant boundary cases.

\begin{definition}[Probability kinematics for infinite partitions and PDFs]
Given: 
\begin{itemize}
\item A random variable $\boldsymbol{\omega}\in\Omega$, where $\Omega$
is a smooth manifold (e.g., the space of $m$ dihedral angles, the $m$-torus $\mathbb{T}^{m}$), equipped with a prior density $\pi(\boldsymbol{\omega})$.
 
\item A random variable $\boldsymbol{\xi}$, defined as a 
many-to-one function of $\boldsymbol{\omega}$. 
For $\boldsymbol{\omega}\in\Omega$, the corresponding realization is written 
$\boldsymbol{\xi}(\boldsymbol{\omega}) \in \Xi$, 
where $\Xi$ is a smooth manifold (e.g., the space of valid pairwise distances 
between $n$ points in $\mathbb{R}^{3}$). 
The prior density of $\boldsymbol{\xi}$ is induced by that of $\boldsymbol{\omega}$ as
\begin{equation}
\pi(\boldsymbol{\xi})
=\int \delta\!\big(\boldsymbol{\xi}(\boldsymbol{\omega})-\boldsymbol{\xi}\big)\,
\pi(\boldsymbol{\omega})\,\mathrm{d}\boldsymbol{\omega}.
\label{eq:prior_pi}
\end{equation}
Each value $\boldsymbol{\xi}\in\Xi$ corresponds to the partition element 
$\{\boldsymbol{\omega}\in\Omega : \boldsymbol{\xi}(\boldsymbol{\omega})=\boldsymbol{\xi}\}$.

\item \textbf{Evidence} as a new density $p(\boldsymbol{\xi})$. 
\item Densities $\pi(\boldsymbol{\xi})$ and $p(\boldsymbol{\xi})$ are
positive on a common support and normalized. We will call $\boldsymbol{\omega}$
and $\boldsymbol{\xi}$ the \textbf{fine-grained} and the \textbf{coarse-grained
or collective} random variables, respectively. 
\item We assume all functions and densities are sufficiently well-behaved
to ensure the existence and measurability of the integrals and conditional
densities involved. 
\end{itemize}
A \textbf{probability kinematics} update defines a posterior density
$p(\boldsymbol{\omega})$ by satisfying the following conditions,
which together ensure the evidence is incorporated while preserving
conditional distributions: 
\begin{enumerate}
\item \textbf{Faithfulness}: The marginal density of $\boldsymbol{\xi}$
matches the evidence,
\begin{equation}
p(\boldsymbol{\xi})=\int p(\boldsymbol{\omega})\delta(\boldsymbol{\xi}(\boldsymbol{\omega})-\boldsymbol{\xi})\textrm{d}\boldsymbol{\omega}.\label{eq:faithfulness}
\end{equation}
\item \textbf{J-Condition}: The conditional density of $\boldsymbol{\omega}$ given $\boldsymbol{\xi}$
is preserved,
\begin{equation}
p(\boldsymbol{\omega}|\boldsymbol{\xi})=\pi(\boldsymbol{\omega}|\boldsymbol{\xi}).\label{eq:j_condition}
\end{equation}
\item \textbf{Probability kinematics update rule}: The posterior $p(\boldsymbol{\omega})$ is obtained
as
\begin{equation}
p(\boldsymbol{\omega}) =\int\pi(\boldsymbol{\omega}\mid\boldsymbol{\xi})p(\boldsymbol{\xi})\textrm{d}\boldsymbol{\xi} \label{eq:pk_integral}.
\end{equation}

\item \textbf{Reference ratio update rule}: Equivalently, after applying Bayes' formula  to (\ref{eq:pk_integral}),
\begin{align}
p(\boldsymbol{\omega}) & =\int\frac{\delta(\boldsymbol{\xi}-\boldsymbol{\xi}(\boldsymbol{\omega}))\pi(\boldsymbol{\omega})}{\pi(\boldsymbol{\xi})}p(\boldsymbol{\xi})\textrm{d}\boldsymbol{\xi}\label{eq:rrm_integral} \\
 & =\frac{p(\boldsymbol{\xi}(\boldsymbol{\omega}))}{\pi(\boldsymbol{\xi}(\boldsymbol{\omega}))}\pi(\boldsymbol{\omega}).\label{eq:rrm}
\end{align}
The Dirac delta function arises from the fact that $p(\boldsymbol{\xi}\mid\boldsymbol{\omega})=\delta(\boldsymbol{\xi}-\boldsymbol{\xi}(\boldsymbol{\omega}))$. Note that the integrand is zero when $\boldsymbol{\xi}\neq\boldsymbol{\xi}(\boldsymbol{\omega})$, and that (\ref{eq:rrm}) follows directly from the J-condition
given by (\ref{eq:j_condition}).
This formulation of PK is useful when the conditional in (\ref{eq:pk_integral}) is unavailable.
\end{enumerate}
\textbf{Consequence}: The posterior $p(\boldsymbol{\omega})$
has marginal $p(\boldsymbol{\xi})$ and preserves the conditional
 $\pi(\boldsymbol{\omega}\mid\boldsymbol{\xi})$. \end{definition}

\subsection{KBPs versus PMFs}

KBPs can be understood as an (approximate) application of PK \citep{Hamelryck_PMF2010,mardia2011lasr_rrm,valentin2014,hamelryck_PK_Book2015}.
Consider a prior distribution over protein structure, $\pi(\boldsymbol{X})$.
Without any loss of generality, we leave out the conditioning on sequence.
We are now given evidence $p(\boldsymbol{D})$ about pairwise distances between a subset of the coordinates, e.g., the $C\beta$ coordinates.
Thus, the coarse-grained variable $\boldsymbol{D}$  induces
a partition on the sample space of the fine-grained variable $\boldsymbol{X}$. These variables correspond to
$\boldsymbol{\xi}$ and $\boldsymbol{\omega}$ in Definition 1, respectively. The PK update following
 (\ref{eq:rrm}) is then given by
\begin{align}
p(\boldsymbol{X}) &=\int\frac{\delta(\boldsymbol{D}(\boldsymbol{X})-\boldsymbol{D})\pi(\boldsymbol{X})}{\pi(\boldsymbol{D})}p(\boldsymbol{D})\textrm{d}\boldsymbol{D} \\
& =\frac{p(\boldsymbol{D}(\boldsymbol{X}))}{\pi(\boldsymbol{D}(\boldsymbol{X}))}\pi(\boldsymbol{X}).\label{eq:KBP-from-PK}
\end{align}
Note that $\boldsymbol{D}$ is a random variable, while $\boldsymbol{D}(\boldsymbol{X})$ is its realization at $\boldsymbol{X}$. A minus-log transform recovers the KBP given by (\ref{eq:Sippl}),
\begin{equation}
\Delta E(\boldsymbol{X})=-\log\left(\frac{p(\boldsymbol{D}(\boldsymbol{X}))}{\pi(\boldsymbol{D}(\boldsymbol{X}))}\right)-\log\left(\pi(\boldsymbol{X})\right),\label{eq:Sippl-from-PK}
\end{equation}
if we assume conditional independence of pairwise distances (see
also Appendix \ref{sec:appendix_approx}). The
second term does not appear in standard KBP expressions, as it is typically implicitly accounted for by Monte Carlo sampling \citep{Rosetta1997}
or scoring of existing  structures \citep{sippl1993recognition}. 

The difference between PK/KBPs and PMFs is now clear: while PMFs amount to
a projection and marginalization of a given Boltzmann distribution of a single physical system as in (\ref{eq:PMF_delta-1}), often formulated relative to a corresponding non-interacting system by subtracting a reference energy as in (\ref{eq:Kirkwood}), PK/KBPs correspond to a generalized Bayesian update
of a prior as in (\ref{eq:pk_integral}-\ref{eq:rrm}). Moreover, PK/KBPs and PMFs differ in producing distributions over fine- ($\boldsymbol{\omega}$) and coarse-grained ($\boldsymbol{\xi}$)
variables, respectively, i.e., 
\begin{align}
\textrm{PMFs:  } p(\boldsymbol{\xi}) &=\int \delta(\boldsymbol{\xi}-\boldsymbol{\xi}(\boldsymbol{\omega}))p(\boldsymbol{\omega})\textrm{d}\boldsymbol{\omega} \\
\textrm{PK/KBPs:  } p(\boldsymbol{\omega}) &=\int\frac{\delta(\boldsymbol{\xi}-\boldsymbol{\xi}(\boldsymbol{\omega}))\pi(\boldsymbol{\omega})}{\pi(\boldsymbol{\xi})}p(\boldsymbol{\xi})\textrm{d}\boldsymbol{\xi}
\end{align}
We now show
that AF1's potential can be understood from the perspective of PK. 

\section{AF1's POTENTIAL RE-EVALUATED\label{sec:AF1's-Potential-Re-evaluated} }

\subsection{AF1's potential as an instance of PK}

In this section, we present a reformulation of the AF1 potential in
probabilistic terms. First, we reinterpret the potential over the
dihedral angles as an empirical prior that is updated with soft evidence on pairwise
distances. Building on previous work on probabilistic models of dihedral
angles \citep{edgoose1998mml,FB52006,Torus08,lennox2009density,Zhao:2010:J-Comput-Biol:20583926,lennox2010dirichlet}
based on directional statistics \citep{mardia2009directional,ley2017modern,pewsey2021recent,mardia2025fisher,mardia2025profile}, this
prior is formulated as 
\begin{align}
\pi(\boldsymbol{\phi},\boldsymbol{\psi};\boldsymbol{\Theta}) & =\prod_{i}\textrm{vM}\left(\phi_{i};\boldsymbol{\Theta}\right)\textrm{vM}\left(\psi_{i};\boldsymbol{\Theta}\right),\label{eq:AF1_prior_angles}
\end{align}
where the parameters are obtained from a deep neural network (NN), $\boldsymbol{\Theta}=\textrm{NN}_{\textrm{Dihedral}}(\boldsymbol{s},\boldsymbol{M})$,
and vM is the von Mises distribution. This circular distribution resembles a Gaussian distribution for large $\kappa$ and is suitable for modeling dihedral angles
in $(-\pi,\pi]$ \citep{mardia2009directional}. The prior over dihedral
angles implies a matching prior over distances,\footnote{We discuss the incorrect assumption of independent pairwise distances in Appendix \ref{sec:appendix_approx}.}
\begin{align}
\pi(\boldsymbol{D};\boldsymbol{\Gamma}) & \propto\prod_{i,j,i\neq j}p\left(D_{i,j};\boldsymbol{\Gamma}_{i,j}\right),\label{eq:AF1_prior_distances}
\end{align}
where $\boldsymbol{\Gamma}=\textrm{NN}_{\textrm{Reference}}(L,\mathbb{I}_{\textrm{Gly}}(\boldsymbol{s}))$. 

The local prior will be updated with soft evidence on distances following the generalized Bayesian
approach of PK. The updated probability distribution
over the distances is,
\begin{align}
p(\boldsymbol{D};\boldsymbol{\Lambda}) & \propto\prod_{i,j,i\neq j}p\left(D_{i,j};\boldsymbol{\Lambda}_{i,j}\right),\label{eq:AF1_update_distances}
\end{align}
where $\boldsymbol{\Lambda}=\textrm{NN}_{\textrm{Distance}}(\boldsymbol{s},\boldsymbol{M})$.
We now have all the elements in place to invoke the machinery of PK. Following PK's reference ratio update given by (\ref{eq:rrm}), which avoids the use of the unavailable conditional used in (\ref{eq:pk_integral}), the posterior is obtained
as
\begin{align}
p(\boldsymbol{\phi},\boldsymbol{\psi};\boldsymbol{\Theta},\boldsymbol{\Gamma},\boldsymbol{\Lambda}) & \propto\frac{p(\boldsymbol{D}(\boldsymbol{\phi},\boldsymbol{\psi});\boldsymbol{\Lambda})}{\pi(\boldsymbol{D}(\boldsymbol{\phi},\boldsymbol{\psi});\boldsymbol{\Gamma})}\pi(\boldsymbol{\phi},\boldsymbol{\psi};\boldsymbol{\Theta}).\label{eq:AF1_final_RRM}
\end{align}

 By construction, AF1 adopts the J-condition induced by a prior over dihedral angles \citep{Torus08,valentin2014}. Why is the PK update needed, if dihedral angles are in principle sufficient to determine the structure? The answer is well known: due to an elbow effect, the dihedral prior is accurate for amino acids that are close together in the protein sequence, but becomes increasingly unreliable for amino acids that are distant \citep{holmes2004some, gong2005building}. This {\em locality} is a well-known limitation of probabilistic models of proteins based on dihedral angles, and is due to the accumulation of small errors and improper covariance representation \citep{arts2024internal}. In AF1, PK provides a minimal and coherent update of the prior, incorporating soft evidence on Euclidean distances while harnessing the prior's useful representation of local structure. The J-condition adopted by AF1 is further discussed in Appendix \ref{sec:appendix_J}. Some practical, approximate aspects of AF1's application of PK, including modeling the pairwise distances as conditionally independent and approximating the reference distribution, are discussed in Appendix \ref{sec:appendix_approx}.

AF1 predictions are obtained  by gradient descent minimization
in the space of the dihedral angles, $\boldsymbol{\phi},\boldsymbol{\psi}$. This approximates a maximum a posteriori (MAP) estimate obtained from
the compositional empirical Bayes model constructed using PK. However, PK allows the formulation of well-defined posterior distributions involving angles, coordinates and distances, as we demonstrate in the next section using a tractable synthetic model.

\section{EXPERIMENTS}\label{sec:experiments}

\begin{figure*}
\centering
\includegraphics[width=0.49\textwidth]{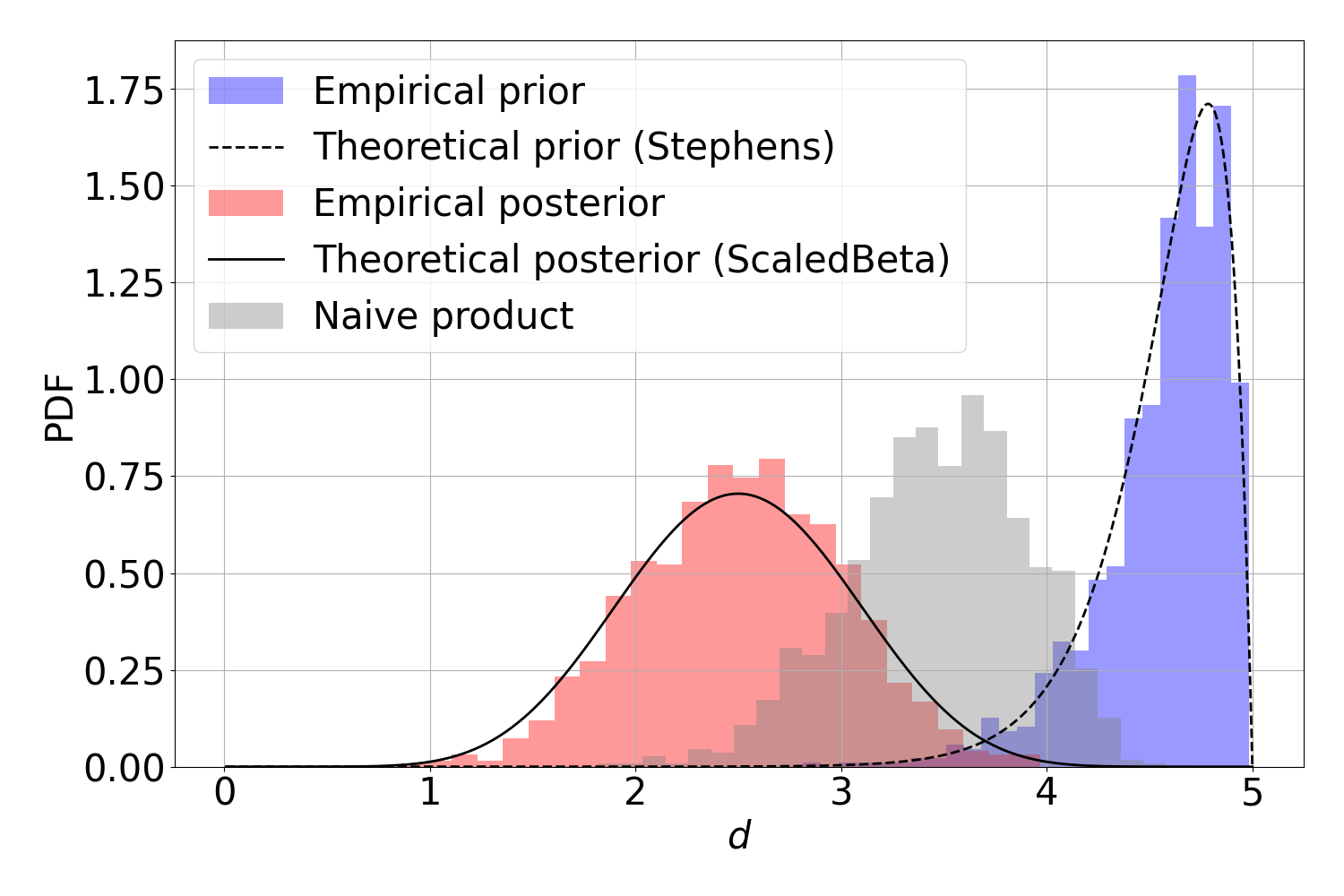}
\hfill
\includegraphics[width=0.49\textwidth]{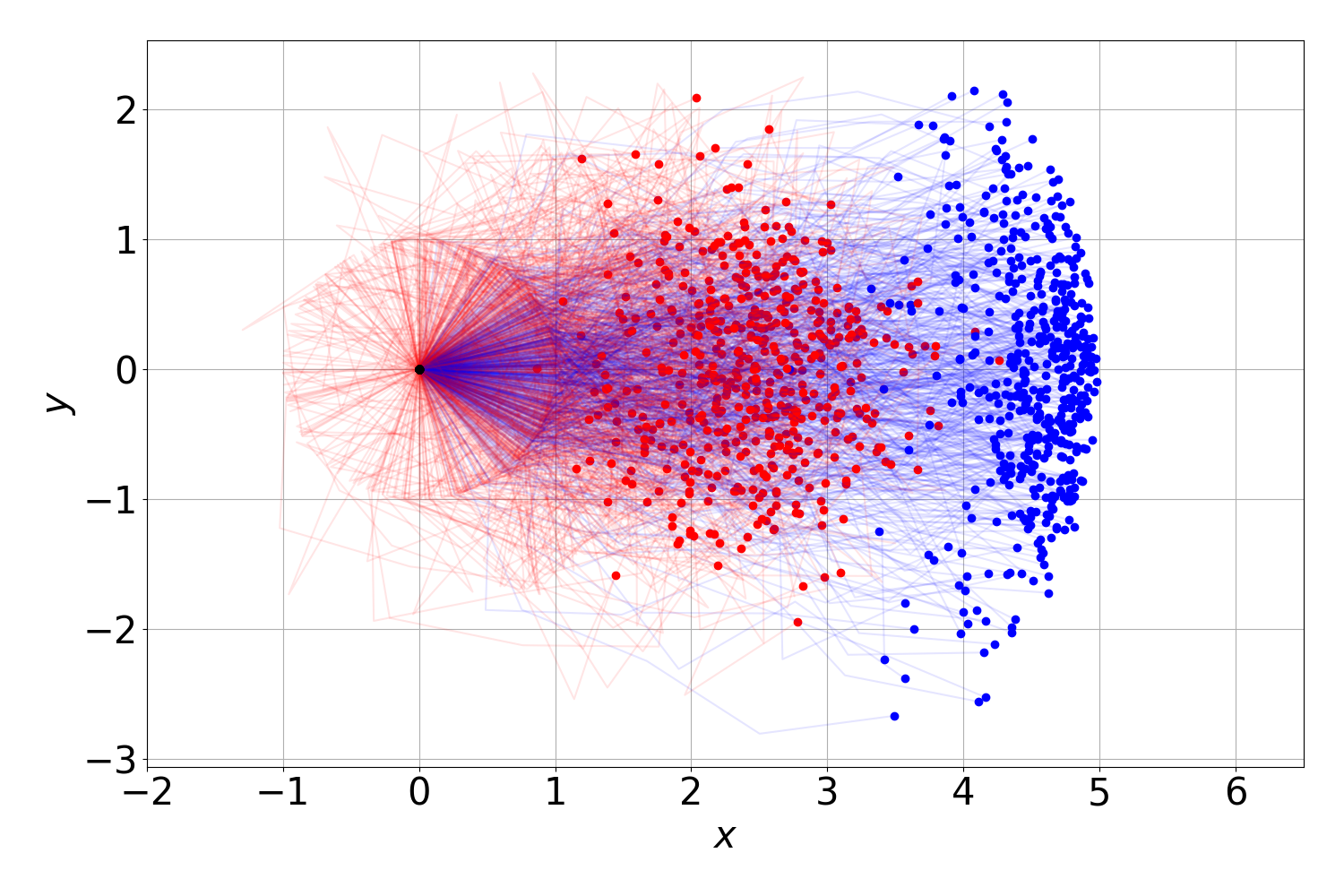}
\par
\caption{\textbf{(left)} Empirical PDFs of the origin-to-end distance,
$d$, for the unmodified VRW prior (with $\mu=0,\kappa=10$, $N=5$; blue histogram), the PK-updated VRW (red histogram), and a
naive product of the $\textrm{ScaledBeta}(\alpha=10,\beta=10,N=5)$ target distribution
with the VRW prior, i.e., without dividing by the reference distribution
(gray histogram). The latter does not fit the target distribution,
clearly showing that the reference distribution
is required. Also shown are the PDFs of the target ScaledBeta distribution
(black line) and of the theoretical distribution for $d$ (Stephens distribution; dashed
line), showing an excellent fit in both cases. \textbf{(right)} 500
2D trajectories, $\boldsymbol{\chi}(\boldsymbol{\theta})$, sampled from
the VRW prior (blue) and the PK-updated distribution (red). A dot
denotes the end point of a random walk. As expected, the endpoints
of the PK-updated distribution lie closer to the origin.\label{fig:VRW}}
\end{figure*}

\paragraph{Motivation}

To demonstrate PK in a way relevant to protein structure prediction,
we formulate a synthetic model that by design has many
crucial properties in common with the AF1 case (summarized in Appendix \ref{sec:appendix_comparison}). For both AF1 and the synthetic model, a generative model that specifies Euclidean coordinates based on an angular distribution is updated using distance evidence.  

Our synthetic model is {\em deliberately minimal} to demonstrate the feasibility and precision of PK updates involving angles, coordinates and distances in a tractable model\footnote{We provide an application to a small 3D peptide model, which requires estimation of the reference distribution, in Appendix \ref{sec:3D_experiment}.}. Our probabilistic framework and synthetic model are thus meant to provide a bedrock for applications to more complex systems in future work.

As prior, we use a von Mises random walk (VRW) in
Euclidean 2D space. As coarse grained variable, we use the length
of the {\em resultant vector}, that is, the distance between the origin and the end of the VRW, which has the advantage that
its distribution is known 
\citep{mardia2009directional,stephens1969tests}. 

The similarities with AF1 include the following elements: (a) both
models make use of the von Mises distribution for a prior over angles, (b) the
angles sampled from the von Mises distributions specify Euclidean
coordinates that can be interpreted as atom positions, (c)
subsequent atomic positions are at a given, fixed distance from each
other, (d) the prior is updated with evidence concerning distances
between a subset of the atoms, (e) the conditional PDF of angles given distances is not available,
requiring the use of PK's reference ratio update given by 
(\ref{eq:rrm}), and (f) the updated
atomic positions are less spatially extended than the prior 
(see also Appendix \ref{sec:appendix_J} and Figure \ref{fig:AF1_3D_samples}). Specifically, as we detail below, angle vector $\boldsymbol{\theta}$, 2D coordinates $\boldsymbol{\chi}$ and distance $d$ of the synthetic model correspond to $(\boldsymbol{\phi}, \boldsymbol{\psi}), \boldsymbol{X}$ and $\boldsymbol{D}$ of AF1, respectively. The correspondences between AF1 and the synthetic model are summarized in Appendix \ref{sec:appendix_comparison}.

\paragraph{Setup}

Let $\boldsymbol{\chi}$ be an $(N+1)\times2$ matrix
of 2-dimensional coordinates, resulting from a VRW with $N$ steps.
The VRW with mean $\mu$ and concentration (i.e., inverse variance)
$\kappa$ is, 
\begin{equation}
\begin{cases}
\boldsymbol{\chi}_{i}=(0,0), & i=0\\
\boldsymbol{\chi}_{i}=\boldsymbol{\chi}_{i-1}+\boldsymbol{v}(\theta_{i}), & 0<i\leq N,
\end{cases}\label{eq:VRW}
\end{equation}
where $\boldsymbol{\theta}=\left\{ \theta_{i}\sim\textrm{vM}(\mu,\kappa)\right\} _{i=1}^{N}$
are angles in $(-\pi,\pi]$ and $\boldsymbol{v}(\theta_{i})$ is the
corresponding 2D unit vector. As in AF1, the coordinates are fully determined
by the angles; as before, we will write $\boldsymbol{\chi}(\boldsymbol{\theta})$ for the realization of $\boldsymbol{\chi}$ at $\boldsymbol{\theta}$.
For this case, an approximate PDF of the Euclidean distance
between start and end of the VRW, i.e., the length of the resultant vector, $d(\boldsymbol{\chi})=\left\Vert \boldsymbol{\chi}_{0}-\boldsymbol{\chi}_{N}\right\Vert$, is given by \cite{stephens1969tests},
\begin{equation}
2N\gamma\left(1-\frac{d}{N}\right)\sim\textrm{Chi}_{N-1}^{2},\label{eq:Stephens}
\end{equation}
where $\frac{1}{\gamma}=\frac{1}{\kappa}+\frac{3}{8\kappa^{2}}$.
For $\kappa \geqslant 4$, the approximation is excellent.
We will write $d\sim\textrm{Stephens}(d;\kappa,N).$ Note that the
maximum value of $d$ is $N$, since each step has unit length. We
define the prior distribution $\pi_{\textrm{VRW}}(\boldsymbol{\theta};\mu,\kappa)$
as the generative model induced by the VRW. The marginal probability
over $d$ for this prior is 
\begin{multline}
\pi_{\textrm{VRW}}(d;\kappa,N) =\textrm{Stephens}(d;\kappa,N) \\
=\int\delta\left(d-d(\boldsymbol{\theta})\right)\pi_{\textrm{VRW}}(\boldsymbol{\theta};\mu,\kappa,N)\textrm{d}\boldsymbol{\theta}.\label{eq:VRW-r}
\end{multline}
Note that $d$ is a random variable, while $d(\boldsymbol{\theta})$ is the realization of $d$ for a specific VRW with angles $\boldsymbol{\theta}$. Now, we will demonstrate how PK allows us to update the prior using
soft information on the distribution of $d$. We wish to modify the
distribution over trajectories such that the marginal distribution
of $d/N$ follows a $\textrm{Beta}(\alpha,\beta)$ distribution\footnote{Note that as $d\in[0,N],$we divide by $N$ to be within the support
of the Beta distribution. } with given parameters $(\alpha,\beta)$ instead of $\textrm{Stephens}(d;\kappa)$.
The {\em target distribution} over $d$ thus becomes after applying the
change of variable formula, 
\begin{multline}
p(d;\alpha,\beta,N)= \\
\frac{1}{N}\frac{1}{B(\alpha,\beta)}\left(\frac{d}{N}\right)^{\alpha-1}\left(1-\frac{d}{N}\right)^{\beta-1},\label{eq:Beta-target}
\end{multline}
where $B(\alpha,\beta)$ is the standard normalization constant of
the Beta distribution and $d \in [0,N]$. We will write $d\sim\textrm{ScaledBeta}(d;\alpha,\beta,N)$.

We can apply PK to update the prior, following
$p(\boldsymbol{\theta})=\int\pi_{\textrm{VRW}}(\boldsymbol{\theta}\mid d)p(d)\textrm{d}\boldsymbol{\theta}$,
but the marginal $\pi_{\textrm{VRW}}(\boldsymbol{\theta}\mid d)$
is not available in a tractable form for a VRW. Therefore, as is the
case for AF1, we use PK's reference ratio update given by (\ref{eq:rrm}),
making use of the fact that the marginal over $d$ according to the
prior is given by (\ref{eq:Stephens}). The PK update has the following form,
\begin{align}
p(\boldsymbol{\theta}) & =\frac{p\left(d(\boldsymbol{\theta})\right)}{\pi_{\textrm{VRW}}\left(d(\boldsymbol{\theta})\right)}\pi_{\textrm{VRW}}(\boldsymbol{\theta}),\label{eq:VRW-RRM}
\end{align}
which, after filling in the concrete PDFs, results in the final model,
\begin{multline}
p(\boldsymbol{\theta};\mu,\kappa,\alpha,\beta,N)  = \\ 
\frac{\textrm{ScaledBeta}\left(d(\boldsymbol{\theta});\alpha,\beta,N\right)}{\textrm{Stephens}\left(d(\boldsymbol{\theta});\kappa,N\right)}\pi_{\textrm{VRW}}(\boldsymbol{\theta};\mu,\kappa,N).\label{eq:VRW-RW-Details}
\end{multline}

We sample from this model using the No-U-Turn Sampler (NUTS) \citep{hoffman2014no}, as implemented in the deep
probabilistic programming language Pyro \citep{bingham2019pyro}, with
default parameters. We use 1000 warm-up and 1000 posterior
samples, taking under one minute on a standard laptop with an Intel i7 CPU. For code availability, see Appendix \ref{sec:code}.

\paragraph{Results and evaluation}

We set $N=5$ and $\mu=0,\kappa=10$
for the VRW, and $\textrm{ScaledBeta}(\alpha=10,\beta=10,N=5)$ as target.
Figure \ref{fig:VRW} visually
confirms that the correct posterior is attained. To further evaluate
the model, we compared the empirical distributions of $d$
to the target distribution using the one-sample Kolmogorov--Smirnov (KS) test \citep{SciPy_2020}, including every fifth NUTS sample to reduce autocorrelation.
Across 10 NUTS simulations, the KS statistic ranges from $0.04$ to $0.08$, with a corresponding median p-value of $0.41$, indicating a good match. 

\paragraph{Ablation experiment}

In contrast and as expected, a naive reweighting approach that omits
the reference distribution produced a much poorer fit, with the KS statistic ranging from $0.62$ to $0.70$, and $p<10^{-70}$ in all cases. The deviation is also very obvious visually (Figure \ref{fig:VRW}, gray histogram). This ablation experiment confirms the importance
of including the reference distribution.

\section{DISCUSSION AND RELATED WORK}\label{sec:discussion}
 
KBPs \citep{sippl_pmf1990} were quantitatively explained by appealing to (\ref{eq:rrm}) by Hamelryck {\em et al.} \citep{Hamelryck_PMF2010,mardia2011lasr_rrm,mardia2012discussion,valentin2014}, and later recognized as an instance of PK \citep{hamelryck_PK_Book2015}. However, these references predate AF1, and are unconnected to deep learning. Moreover, the connection between KBPs and PK, while formally sound, has remained virtually unexplored within the deep learning paradigm applied to biomolecular structure. To our knowledge, our work is the first to analyze AF1 as an instance of PK.

Although PK is a well-defined concept in probability theory \citep{jeffrey1965logic,diaconis1982updating,jeffrey2004subjective,chan2005revision,meehan2020jeffrey}, as far as we know, it has seen almost no explicit adoption in deep models: \cite{yu2020bayesian} applies PK to training Bayesian neural networks with soft evidence. \cite{munk2023uncertain} use PK to incorporate soft evidence in probabilistic models and stochastic simulators. 

This work connects statistical mechanics, probabilistic modeling, and deep learning, which are increasingly converging in the field of protein structure prediction \citep{force-field-ML20232, jing2025ai, olsson2026generative}. The success of AF1 and its roots in PK hint at a general approach to formulate deep models that are compositional \citep{ghahramani2015probabilistic,bingham2019pyro,wood2014new}. This approach consists of updating an empirical Bayes prior concerning a fine-grained variable with soft evidence on a coarse-grained variable, both obtained from deep models.  We point out that PK is agnostic regarding the nature of its components, bringing a wide range of deep generative models potentially within its scope \citep{tomczak2024deep}.

Finally, probability kinematics has been connected to \emph{credal set theory}, where uncertainty is represented by closed, convex sets of probability distributions rather than a single posterior  \citep{marchetti2018imaginary, caprio2023dynamic, caprio2023ergodic}, providing opportunities for formulating deep generative models that can explicitly accommodate imprecise, partially conflicting or even inconsistent information.  

\section{CONCLUSIONS}
We interpret AF1 as an instance of PK and demonstrated the probabilistic framework using a synthetic model with qualitative evaluation. Beyond clarifying the foundations of a paradigm-shifting method, our results imply that PK offers a principled foundation for building compositional deep probabilistic models based on generalized Bayesian updating \citep{ghahramani2015probabilistic,bingham2019pyro,wood2014new}.

\section{CODE AVAILABILITY}\label{sec:code}
Code for reproducing the synthetic von Mises random walk experiment in Section~\ref{sec:experiments} is available at:
\url{https://github.com/thamelry/AlphaFoldPK}.

\section{ACKNOWLEDGMENTS}
We thank Wouter Boomsma, Jesper Ferkinghoff-Borg, Jes Frellsen, Carsten Hartmann, John T. Kent, Ola Rønning, Manfred Sippl and Sandy Zabell for discussions and comments, and an anonymous reviewer for pointing out the connection to credal sets. We acknowledge funding from the VILLUM Experiment Programme (grant 50240) and the Independent Research Fund Denmark (grant 10.46540/5244-00031B).

\newpage

\bibliographystyle{abbrvnat}
\bibliography{af}

@article{jing2025ai,
  title={{AI}-based Methods for Simulating, Sampling, and Predicting Protein Ensembles},
  author={Jing, Bowen and Berger, Bonnie and Jaakkola, Tommi},
  journal={arXiv:2509.17224},
  year={2025}
}

@article{olsson2026generative,
  title={Generative molecular dynamics},
  author={Olsson, Simon},
  journal={Current Opinion in Structural Biology},
  volume={96},
  pages={103213},
  year={2026}
}

@article{marchetti2018imaginary,
  title={Imaginary kinematics},
  author={Marchetti, Sabina and Antonucci, Alessandro},
  journal={arXiv:1808.00329},
  year={2018}
}

@inproceedings{caprio2023dynamic,
  title={Dynamic precise and imprecise probability kinematics},
  author={Caprio, Michele and Gong, Ruobin},
  booktitle={International Symposium on Imprecise Probability: Theories and Applications (ISIPTA)},
  pages={72--83},
  year={2023}
}

@article{caprio2023ergodic,
  title={Ergodic theorems for dynamic imprecise probability kinematics},
  author={Caprio, Michele and Mukherjee, Sayan},
  journal={International Journal of Approximate Reasoning},
  volume={152},
  pages={325--343},
  year={2023}
}

@article{zhou2002distance,
  title={Distance-scaled, finite ideal-gas reference state improves structure-derived potentials of mean force for structure selection and stability prediction},
  author={Zhou, Hongyi and Zhou, Yaoqi},
  journal={Protein Science},
  volume={11},
  pages={2714--2726},
  year={2002},
  publisher={Wiley Online Library}
}

@article{shen2006statistical,
  title={Statistical potential for assessment and prediction of protein structures},
  author={Shen, Min-Yi and Sali, Andrej},
  journal={Protein Science},
  volume={15},
  pages={2507--2524},
  year={2006},
  publisher={Wiley Online Library}
}

@article{zhang2010novel,
  title={A novel side-chain orientation dependent potential derived from random-walk reference state for protein fold selection and structure prediction},
  author={Zhang, Jian and Zhang, Yang},
  journal={{PLoS ONE}},
  volume={5},
  pages={e15386},
  year={2010},
  publisher={Public Library of Science San Francisco, USA}
}

@article{lu2008opus,
  title={{OPUS-PSP}: {A}n orientation-dependent statistical all-atom potential derived from side-chain packing},
  author={Lu, Mingyang and Dousis, Athanasios D and Ma, Jianpeng},
  journal={Journal of Molecular Biology},
  volume={376},
  pages={288--301},
  year={2008},
  publisher={Elsevier}
}

@article{parsons2005practical,
  title={Practical conversion from torsion space to {C}artesian space for in silico protein synthesis},
  author={Parsons, Jerod and Holmes, J Bradley and Rojas, J Maurice and Tsai, Jerry and Strauss, Charlie EM},
  journal={Journal of Computational Chemistry},
  volume={26},
  pages={1063--1068},
  year={2005}
}

@article{alquraishi2019parallelized,
  title={Parallelized natural extension reference frame: {P}arallelized conversion from internal to {C}artesian coordinates},
  author={AlQuraishi, Mohammed},
  journal={Journal of Computational Chemistry},
  volume={40},
  pages={885--892},
  year={2019}
}

@article{holmes2004some,
  title={Some fundamental aspects of building protein structures from fragment libraries},
  author={Holmes, J Bradley and Tsai, Jerry},
  journal={Protein Science},
  volume={13},
  pages={1636--1650},
  year={2004},
  publisher={Wiley Online Library}
}

@article{gong2005building,
  title={Building native protein conformation from highly approximate backbone torsion angles},
  author={Gong, Haipeng and Fleming, Patrick J and Rose, George D},
  journal={Proceedings of the National Academy of Sciences},
  volume={102},
  pages={16227--16232},
  year={2005},
  publisher={National Academy of Sciences}
}

@book{tomczak2024deep,
  title={Deep generative modeling},
  author={Tomczak, Jakub M},
  edition={2nd},
  publisher={Springer},
  year={2024}
}

@article{pohl1971empirical,
  title={Empirical protein energy maps},
  author={Pohl, F. M.},
  journal={Nature New Biology},
  volume={234},
  pages={277--279},
  year={1971}
}

@article{mardia2012discussion,
  author    = {K. V. Mardia and T. Hamelryck},
  title     = {On ‘{C}onstructing summary statistics for approximate {B}ayesian computation: {S}emi-automatic approximate {B}ayesian computation’ by {P. Fearnhead and D. Prangle}},
  journal   = {Journal of the Royal Statistical Society: Series B (Statistical Methodology)},
  year      = {2012},
  volume    = {74},
  pages     = {462--463},
  note      = {Discussion},
}

@inproceedings{mardia2011lasr_rrm,
  title={A statistical view on the reference ratio method},
  author={Mardia, Kanti V and Frellsen, Jes and Borg, Mikael and Ferkinghoff-Borg, Jesper and Hamelryck, Thomas},
  booktitle={LASR 2011 Proceedings},
  year = {2011},
  pages={55--61},
  publisher = {Leeds University Press}
}

@article{mardia2025profile,
  title={Profile: {K}anti {M}ardia}, 
  author={Mardia, Kanti V},
  journal={Significance},
  volume={22},
  pages={36--38},
  year={2025}
}

@article{mardia2025fisher,
  title={Fisher’s legacy of directional statistics, and beyond to statistics on manifolds},
  author={Mardia, Kanti V},
  journal={Journal of Multivariate Analysis},
  volume={207},
  pages={105404},
  year={2025}
}

@article{pewsey2021recent,
  title={Recent advances in directional statistics},
  author={Pewsey, Arthur and Garc{\'\i}a-Portugu{\'e}s, Eduardo},
  journal={TEST},
  volume={30},
  pages={1--58},
  year={2021},
  publisher={Springer}
}

@book{ley2017modern,
  title={Modern directional statistics},
  author={Ley, Christophe and Verdebout, Thomas},
  year={2017},
  publisher={Chapman and Hall/CRC}
}

@inproceedings{wood2014new,
  title={A new approach to probabilistic programming inference},
  author={Wood, Frank and Meent, Jan Willem and Mansinghka, Vikash},
  booktitle={Artificial Intelligence and Statistics (AISTATS)},
  pages={1024--1032},
  year={2014},
  organization={PMLR}
}

@article{kingma2019introduction,
  title={An introduction to variational autoencoders},
  author={Kingma, Diederik P and Welling, Max},
  journal={Foundations and Trends in Machine Learning},
  volume={12},
  pages={307--392},
  year={2019}
}

@article{blei2017variational,
  title={Variational inference: {A} review for statisticians},
  author={Blei, David M and Kucukelbir, Alp and McAuliffe, Jon D},
  journal={Journal of the American Statistical Association},
  volume={112},
  pages={859--877},
  year={2017}
}

@article{ramachandran1963stereochemistry,
  title={Stereochemistry of polypeptide chain configurations},
  author={Ramachandran, GN and Ramakrishnan, C and Sasisekharan, V},
  journal={Journal of Molecular Biology},
  volume={7},
  pages={95--99},
  year={1963}
}

@article{ghahramani2015probabilistic,
  title={Probabilistic machine learning and artificial intelligence},
  author={Ghahramani, Zoubin},
  journal={Nature},
  volume={521},
  pages={452--459},
  year={2015}
}

@book{zuckerman2010statistical,
  title={Statistical physics of biomolecules: {A}n introduction},
  author={Zuckerman, Daniel M},
  year={2010},
  publisher={CRC press}
}

@ARTICLE{SciPy_2020,
  author  = {Virtanen, Pauli and Gommers, Ralf and Oliphant, Travis E. and
            Haberland, Matt and Reddy, Tyler and Cournapeau, David and
            Burovski, Evgeni and Peterson, Pearu and Weckesser, Warren and
            Bright, Jonathan and others},
  title   = {{SciPy} 1.0: {F}undamental Algorithms for Scientific
            Computing in {P}ython},
  journal = {Nature Methods},
  year    = {2020},
  volume  = {17},
  pages   = {261--272}
}

@article{sippl1995knowledge,
  title={Knowledge-based potentials for proteins},
  author={Sippl, Manfred J},
  journal={Current Opinion in Structural Biology},
  volume={5},
  pages={229--235},
  year={1995}
}

@article{sippl1993boltzmann,
  title={Boltzmann's principle, knowledge-based mean fields and protein folding. {A}n approach to the computational determination of protein structures},
  author={Sippl, Manfred J},
  journal={Journal of Computer-Aided Molecular Design},
  volume={7},
  pages={473--501},
  year={1993}
}

@article{sippl1996helmholtz,
  title={Helmholtz free energies of atom pair interactions in proteins},
  author={Sippl, Manfred J and Ortner, Maria and Jaritz, Markus and Lackner, Peter and Fl{\"o}ckner, Hannes},
  journal={Folding and Design},
  volume={1},
  pages={289--298},
  year={1996}
}

@book{hamelryck2012bayesian,
  title={{Bayesian methods in structural bioinformatics}},
  editor={Hamelryck, Thomas and Mardia, Kanti V. and Ferkinghoff-Borg, Jesper},
  year={2012},
  publisher={Springer}
}

@article{chan2005revision,
  title={On the revision of probabilistic beliefs using uncertain evidence},
  author={Chan, Hei and Darwiche, Adnan},
  journal={Artificial Intelligence},
  volume={163},
  pages={67--90},
  year={2005}
}

@inproceedings{munk2023uncertain,
  title={Uncertain evidence in probabilistic models and stochastic simulators},
  author={Munk, Andreas and Mead, Alexander and Wood, Frank},
  booktitle={International Conference on Machine Learning (ICML)},
  pages={25486--25500},
  year={2023}
}

@article{kabsch1976,
  title={A solution for the best rotation to relate two sets of vectors},
  author={Kabsch, Wolfgang},
  journal={Acta Crystallographica A},
  volume={32},
  pages={922--923},
  year={1976}
}

@unpublished{PyMOL,
    Title = {The {PyMOL} Molecular Graphics System, {V}ersion 3.0.0},
    Author = {{Schr\"odinger, LLC}},
    year = {2015},
    Note = {Available at \url{https://pymol.org/}}
}

@article{force-field-ML20232,
  title={Machine learned coarse-grained protein force-fields: {A}re we there yet?},
  author={Durumeric, Aleksander EP and Charron, Nicholas E and Templeton, Clark and Musil, F{\'e}lix and Bonneau, Klara and Pasos-Trejo, Aldo S and Chen, Yaoyi and Kelkar, Atharva and No{\'e}, Frank and Clementi, Cecilia},
  journal={Current Opinion in Structural Biology},
  volume={79},
  pages={102533},
  year={2023}
}

@article{yu2020bayesian,
  title={Bayesian Neural Networks with Soft Evidence},
  author={Yu, Edward},
  journal={preprint arXiv:2010.09570},
  year={2020}
}

@article{af1_nature_2020,
  title={Improved protein structure prediction using potentials from deep learning},
  author={Senior, Andrew W and Evans, Richard and Jumper, John and Kirkpatrick, James and Sifre, Laurent and Green, Tim and Qin, Chongli and {\v{Z}}{\'\i}dek, Augustin and Nelson, Alexander WR and Bridgland, Alex and others},
  journal={Nature},
  volume={577},
  pages={706--710},
  year={2020}
}

@article{chakravarty2024alphafold,
  title={{AlphaFold} predictions of fold-switched conformations are driven by structure memorization},
  author={Chakravarty, Devlina and Schafer, Joseph W and Chen, Ethan A and Thole, Joseph F and Ronish, Leslie A and Lee, Myeongsang and Porter, Lauren L},
  journal={Nature Communications},
  volume={15},
  pages={7296},
  year={2024}
}

@article{bowman2024alphafold,
  title={{AlphaFold} and protein folding: {Not} dead yet! {T}he frontier is conformational ensembles},
  author={Bowman, Gregory R},
  journal={Annual Review of Biomedical Data Science},
  volume={7},
  pages={51--57},
  year={2024},
  publisher={Annual Reviews}
}

@article{jumper2021highly,
  title={Highly accurate protein structure prediction with {AlphaFold}},
  author={Jumper, John and Evans, Richard and Pritzel, Alexander and Green, Tim and Figurnov, Michael and Ronneberger, Olaf and Tunyasuvunakool, Kathryn and Bates, Russ and {\v{Z}}{\'\i}dek, Augustin and Potapenko, Anna and others},
  journal={Nature},
  volume={596},
  pages={583--589},
  year={2021},
  publisher={Nature Publishing Group}
}

@article{abramson2024accurate,
  title={Accurate structure prediction of biomolecular interactions with {AlphaFold3}},
  author={Abramson, Josh and Adler, Jonas and Dunger, Jack and Evans, Richard and Green, Tim and Pritzel, Alexander and Ronneberger, Olaf and Willmore, Lindsay and Ballard, Andrew J and Bambrick, Joshua and others},
  journal={Nature},
  volume={630},
  pages={493--500},
  year={2024},
  publisher={Nature Publishing Group UK London}
}

@article{stephens1969tests,
  title={Tests for the von {M}ises distribution},
  author={Stephens, M A},
  journal={Biometrika},
  volume={56},
  pages={149--160},
  year={1969}
}

@article{callaway2024chemistry,
  title={Chemistry {N}obel goes to developers of {AlphaFold AI} that predicts protein structures},
  author={Callaway, Ewen},
  journal={Nature},
  volume={634},
  pages={525--526},
  year={2024}
}

@article{chakravarty2025proteins,
  title={Proteins with alternative folds reveal blind spots in {AlphaFold}-based protein structure prediction},
  author={Chakravarty, Devlina and Lee, Myeongsang and Porter, Lauren L},
  journal={Current Opinion in Structural Biology},
  volume={90},
  pages={102973},
  year={2025},
  publisher={Elsevier}
}

@article{gavalda2025gradations,
  title={Gradations in protein dynamics captured by experimental {NMR} are not well represented by {AlphaFold2} models and other computational metrics},
  author={Gavalda-Garcia, Jose and Dixit, Bhawna and D{\'\i}az, Adri{\'a}n and Ghysels, An and Vranken, Wim},
  journal={Journal of Molecular Biology},
  volume={437},
  pages={168900},
  year={2025},
  publisher={Elsevier}
}

@article{ramasamy2025assessing,
  title={Assessing the relation between protein phosphorylation, {AlphaFold3 models} and conformational variability},
  author={Ramasamy, Pathmanaban and Zuallaert, Jasper and Martens, Lennart and Vranken, Wim F},
  journal={preprint bioRxiv},
  year={2025}
}

@article{agarwal2024power,
  title={The power and pitfalls of {AlphaFold2} for structure prediction beyond rigid globular proteins},
  author={Agarwal, Vinayak and McShan, Andrew C},
  journal={Nature Chemical Biology},
  volume={20},
  pages={950--959},
  year={2024},
  publisher={Nature Publishing Group US New York}
}

@article{AF12019protein,
  title={Protein structure prediction using multiple deep neural networks in the 13th {Critical Assessment of Protein Structure Prediction (CASP13})},
  author={Senior, Andrew W and Evans, Richard and Jumper, John and Kirkpatrick, James and Sifre, Laurent and Green, Tim and Qin, Chongli and {\v{Z}}{\'\i}dek, Augustin and Nelson, Alexander WR and Bridgland, Alex and others},
  journal={{Proteins: Structure, Function, and Bioinformatics}},
  volume={87},
  pages={1141--1148},
  year={2019}
}

@article{armendt1980there,
  title={Is there a {D}utch book argument for probability kinematics?},
  author={Armendt, Brad},
  journal={Philosophy of Science},
  volume={47},
  pages={583--588},
  year={1980},
  publisher={Cambridge University Press}
}

@article{diaconis1982updating,
  title={Updating subjective probability},
  author={Diaconis, Persi and Zabell, Sandy L},
  journal={{Journal of the American Statistical Association}},
  volume={77},
  pages={822--830},
  year={1982},
  publisher={Taylor \& Francis}
}

@article{meehan2020jeffrey,
  title={Jeffrey meets {K}olmogorov: {A} general theory of conditioning},
  author={Meehan, Alexander and Zhang, Snow},
  journal={{Journal of Philosophical Logic}},
  volume={49},
  pages={941--979},
  year={2020},
  publisher={Springer}
}

@article{bingham2019pyro,
  title={Pyro: {D}eep universal probabilistic programming},
  author={Bingham, Eli and Chen, Jonathan P and Jankowiak, Martin and Obermeyer, Fritz and Pradhan, Neeraj and Karaletsos, Theofanis and Singh, Rohit and Szerlip, Paul and Horsfall, Paul and Goodman, Noah D},
  journal={Journal of Machine Learning Research},
  volume={20},
  pages={1--6},
  year={2019}
}

@article{hoffman2014no,
  title={The {No-U-Turn} sampler: {A}daptively setting path lengths in {Hamiltonian Monte Carlo}.},
  author={Hoffman, Matthew D and Gelman, Andrew},
  journal={Journal of Machine Learning Research},
  volume={15},
  pages={1593--1623},
  year={2014}
}

@inproceedings{edgoose1998mml,
  title={An {MML} classification of protein structure that knows about angles and sequence.},
  author={Edgoose, T and Allison, L and Dowe, DL},
  booktitle={Pacific Symposium on Biocomputing},
  pages={585--596},
  year={1998}
}

@article{lennox2010dirichlet,
  title={A {D}irichlet process mixture of hidden {M}arkov models for protein structure prediction},
  author={Lennox, Kristin P and Dahl, David B and Vannucci, Marina and Day, Ryan and Tsai, Jerry W},
  journal={Annals of Applied Statistics},
  volume={4},
  pages={916},
  year={2010}
}

@article{lennox2009density,
  title={Density estimation for protein conformation angles using a bivariate von {M}ises distribution and {B}ayesian nonparametrics},
  author={Lennox, Kristin P and Dahl, David B and Vannucci, Marina and Tsai, Jerry W},
  journal={Journal of the American Statistical Association},
  volume={104},
  pages={586--596},
  year={2009},
  publisher={Taylor \& Francis}
}

@book{mardia2009directional,
  title={Directional statistics},
  author={Mardia, Kanti V and Jupp, Peter E},
  year={2009},
  publisher={John Wiley \& Sons}
}

@book{jeffrey1957contributions,
  title={Contributions to the theory of inductive probability},
  author={Jeffrey, Richard C},
  year={1957},
  publisher={PhD thesis, Princeton University}
}

@book{jeffrey1965logic,
  title={The logic of decision},
  author={Jeffrey, Richard C},
  year={1990},
  edition = {2nd},
  publisher={The University of Chicago Press},
  note      = {Originally published in 1965 by McGraw-Hill}
}

@book{jeffrey2004subjective,
  title={Subjective probability: {T}he real thing},
  author={Jeffrey, Richard C},
  year={2004},
  publisher={Cambridge University Press}
}

@article{dill2012protein,
  title={The Protein-Folding Problem, 50 Years On},
  author={Dill, K.A. and MacCallum, J.L.},
  journal={Science},
  volume={338},
  pages={1042--1046},
  year={2012}
}

@article{kirkwood1935,
        author = {John G. Kirkwood},
        title = {Statistical Mechanics of Fluid Mixtures},
        publisher = {AIP},
        year = {1935},
        journal = {Journal of Chemical Physics},
        volume = {3},
        pages = {300-313},
}

@Article{sippl_pmf1990,
   Author="Sippl, M. J. ",
   Title="{{C}alculation of conformational ensembles from potentials of mean force. {A}n approach to the knowledge-based prediction of local structures in globular proteins}",
   Journal="Journal of Molecular Biology",
   Year="1990",
   Volume="213",
   Pages="859--883",
}

@article{Torus08,
  Author         = {Boomsma, W. and Mardia, KV. and Taylor, CC. and
  Ferkinghoff-Borg, J. and Krogh, A. and Hamelryck, T.},
  Title          = {A generative, probabilistic model of local protein
  structure.},
  Journal        = {Proceedings of the National Academy USA},
  Volume         = {105},
  Pages          = {8932--8937},
  year           = {2008}
}

@article{FB52006,
  Author         = {Hamelryck, T. and Kent, JT. and Krogh, A.},
  Title          = {Sampling realistic protein conformations using local structural bias},
  Journal        = {PLoS Computational Biology},
  Volume         = {2},
  Number         = {9},
  Pages          = {e131},
  year           = {2006}
}

@Article{Zhao:2010:J-Comput-Biol:20583926,
author = "Zhao, F. and Peng, J. and Debartolo, J. and Freed, K.F. and
Sosnick, T.R. and Xu, J.",
title = {A probabilistic and continuous model of protein conformational
space for template-free modeling},
journal = "Journal of Computional Biology",
year = "2010",
volume = "17",
pages = "783-798",
}

@article{finkelstein1995protein,
  title={Why do protein architectures have {B}oltzmann-like statistics?},
  author={Finkelstein, Alexei V and Badretdinov, Azat Ya and Gutin, Alexander M},
  journal={{Proteins: Structure, Function, and Bioinformatics}},
  volume={23},
  pages={142--150},
  year={1995}
}

@Article{Rosetta1997,
   Author="Simons, K. T.  and Kooperberg, C.  and Huang, E.  and Baker, D. ",
   Title="{{A}ssembly of protein tertiary structures from fragments with similar local sequences using simulated annealing and {B}ayesian scoring functions}",
   Journal="Journal of Molecular Biology",
   Year="1997",
   Volume="268",
   Pages="209--225",
}

@article{hartmann_pmf2011,
  title={On two possible definitions of the free energy for collective variables},
  author={Hartmann, Carsten and Latorre, Juan C and Ciccotti, Giovanni},
  journal={European Physical Journal Special Topics},
  volume={200},
  pages={73--89},
  year={2011}
}

@article{lopez2025severe,
  title={Severe deviation in protein fold prediction by advanced {AI}: {A} case study},
  author={L{\'o}pez-Sagaseta, Jacinto and Urdiciain, Alejandro},
  journal={{Scientific Reports}},
  volume={15},
  pages={4778},
  year={2025}
}

@Article{pmid8609636,
   Author="Thomas, P. D.  and Dill, K. A. ",
   Title="{Statistical potentials extracted from protein structures: {H}ow accurate are they?}",
   Journal="{Journal of Molecular Biology}",
   Year="1996",
   Volume="257",
   Pages="457--469",
}

@article{ben-naim:3698,
        author = {A. Ben-Naim},
        title = {{Statistical potentials extracted from protein
structures: {A}re these meaningful potentials?}},
        year = {1997},
        journal = {Journal of Chemical Physics},
        volume = {107},
        pages = {3698-3706},
}

@Article{pmid9526121,
   Author="Koppensteiner, W. A.  and Sippl, M. J. ",
   Title="{{K}nowledge-based potentials -- {B}ack to the roots}",
   Journal={{Biochemistry (Moscow)}},
   Year="1998",
   Volume="63",
   Pages="247--252",
}

@Book{chandler1987,
        author = {Chandler, D.},
        title = {Introduction to modern statistical mechanics},
        publisher = {Oxford University Press},
        year = {1987},
}

@article{sippl1993recognition,
  title={Recognition of errors in three-dimensional structures of proteins},
  author={Sippl, Manfred J},
  journal={{Proteins: Structure, Function, and Bioinformatics}},
  volume={17},
  pages={355--362},
  year={1993}
}

@article{tanaka1976medium,
  title={Medium-and long-range interaction parameters between amino acids for predicting three-dimensional structures of proteins},
  author={Tanaka, Seiji and Scheraga, Harold A},
  journal={Macromolecules},
  volume={9},
  pages={945--950},
  year={1976}
}

@article{miyazawa1985estimation,
  title={Estimation of effective interresidue contact energies from protein crystal structures: {Q}uasi-chemical approximation},
  author={Miyazawa, Sanzo and Jernigan, Robert L},
  journal={Macromolecules},
  volume={18},
  pages={534--552},
  year={1985}
}

@article{Hamelryck_PMF2010,
    author={Hamelryck, T. and Borg, M. and Paluszewski, M. and Paulsen, J. and Frellsen, J. and Andreetta, C. and Boomsma, W. and Bottaro, S. and Ferkinghoff-Borg, J.},
    year={2010},
    title={Potentials of mean force for protein structure prediction vindicated, formalized and generalized},
    journal={PLoS ONE},
    volume={5},
    pages={e13714}
}

@article{valentin2014,
  title={Formulation of probabilistic models of protein structure in atomic detail using the reference ratio method},
  author={Valentin, Jan B and Andreetta, Christian and Boomsma, Wouter and Bottaro, Sandro and Ferkinghoff-Borg, Jesper and Frellsen, Jes and Mardia, Kanti V and Tian, Pengfei and Hamelryck, Thomas},
  journal={{Proteins: Structure, Function, and Bioinformatics}},
  volume={82},
  pages={288--299},
  year={2014}
}

@InCollection{hamelryck_PK_Book2015,
  editor={Dryden, IL. and Kent, JT.},
  title={Proteins, physics and probability kinematics: {A} {B}ayesian formulation of the protein folding problem},
  author={Hamelryck, Thomas and Boomsma, Wouter and Ferkinghoff-Borg, Jesper and Foldager, Jesper and Frellsen, Jes and Haslett, John and Theobald, Douglas},
  booktitle={Geometry Driven Statistics},
  pages={356--376},
  year={2015},
  publisher={Wiley}
}

@inproceedings{clifford1985distances,
  title={Distances in {G}aussian point sets},
  author={Clifford, Peter and Green, NJB},
  booktitle={{Mathematical Proceedings of the Cambridge Philosophical Society}},
  volume={97},
  pages={515--524},
  year={1985},
  organization={Cambridge University Press}
}

@article{arts2024internal,
  title={Internal-Coordinate Density Modelling of Protein Structure: {C}ovariance Matters},
  author={Arts, Marloes and Frellsen, Jes and Boomsma, Wouter},
  journal={Transactions on Machine Learning Research},
  year={2024}
}

@book{Dryden2016,
    author = {Dryden, I. L. and Mardia, K. V.},
    publisher = {Wiley},
    title = {Statistical shape analysis with applications in {R}, Second Edition},
    year = {2016}
}

\newpage

\ifpreprintclass
\else
\section*{Checklist}

\begin{enumerate}

  \item For all models and algorithms presented, check if you include:
  \begin{enumerate}
    \item A clear description of the mathematical setting, assumptions, algorithm, and/or model. [Yes]
    \item An analysis of the properties and complexity (time, space, sample size) of any algorithm. [Yes]
    \item (Optional) Anonymized source code, with specification of all dependencies, including external libraries. [Yes]
  \end{enumerate}

  \item For any theoretical claim, check if you include:
  \begin{enumerate}
    \item Statements of the full set of assumptions of all theoretical results. [Yes]
    \item Complete proofs of all theoretical results. [Yes]
    \item Clear explanations of any assumptions. [Yes]
  \end{enumerate}

  \item For all figures and tables that present empirical results, check if you include:
  \begin{enumerate}
    \item The code, data, and instructions needed to reproduce the main experimental results (either in the supplemental material or as a URL). [Yes]
    \item All the training details (e.g., data splits, hyperparameters, how they were chosen). [Yes]
    \item A clear definition of the specific measure or statistics and error bars (e.g., with respect to the random seed after running experiments multiple times). [Yes]
    \item A description of the computing infrastructure used. (e.g., type of GPUs, internal cluster, or cloud provider). [Yes]
  \end{enumerate}

  \item If you are using existing assets (e.g., code, data, models) or curating/releasing new assets, check if you include:
  \begin{enumerate}
    \item Citations of the creator, if your work uses existing assets. [Yes]
    \item The license information of the assets, if applicable. [Not Applicable]
    \item New assets either in the supplemental material or as a URL, if applicable. [Not Applicable]
    \item Information about consent from data providers/curators. [Not Applicable]
    \item Discussion of sensible content if applicable, e.g., personally identifiable information or offensive content. [Not Applicable]
  \end{enumerate}

  \item If you used crowdsourcing or conducted research with human subjects, check if you include:
  \begin{enumerate}
    \item The full text of instructions given to participants and screenshots. [Not Applicable]
    \item Descriptions of potential participant risks, with links to Institutional Review Board (IRB) approvals if applicable. [Not Applicable]
    \item The estimated hourly wage paid to participants and the total amount spent on participant compensation. [Not Applicable]
  \end{enumerate}

\end{enumerate}
\fi

\onecolumn
\appendix
\section*{APPENDIX}
\section{NOTATION AND ABBREVIATIONS}\label{sec:appendix_notation}

\begin{table}[h!]
\centering
\begin{tabular}{ll}
\hline
\textbf{Symbol / Abbreviation} & \textbf{Meaning (and shape of arrays)} \\
\hline
AF1 & AlphaFold version 1 \\
KBP & Knowledge Based Potential  \\
MAP & Maximum A Posteriori estimate \\
MSA & Multiple Sequence Alignment \\
$\textrm{NN}$ & Neural Network \\
$\mathcal{N}(\mu, \sigma^2)$ & Normal distribution with mean $\mu$ and variance $\sigma^2$ \\
PK & Probability Kinematics (Jeffrey conditioning) \\
PMF & Potential of Mean Force \\
vM$(\phi; \mu, \kappa)$ & von Mises distribution, PDF with support $(-\pi,\pi]$  \\
VRW & von Mises Random Walk \\
$\pi(\cdot)$ & Prior PDF \\
\hline

AF1 & \\
\hline
$L$ & Length (number of amino acids) of a protein \\
$\boldsymbol{s}$ & Protein sequence, one-hot encoding, shape $L \times 20$ \\
$\boldsymbol{M}$ & MSA, shape $M \times L \times 20$ for $M$ sequences\\
$\mathbb{I}_{\textrm{Gly}}(\boldsymbol{s})$ & Glycines in protein sequence, one-hot encoding, shape $L$  \\
$\boldsymbol{X}$ & Coordinates of protein backbone, shape $L\times 4\times 3$ \\
$\boldsymbol{X}(\boldsymbol{\phi}, \boldsymbol{\psi})$ & Realization of $\boldsymbol{X}$ for given   dihedral angles \\ 
$\boldsymbol{\phi}, \boldsymbol{\psi}$ & Protein dihedral angles in $(-\pi,\pi]$, shape $L$ each \\
$\boldsymbol{D}$ & C$\beta$-C$\beta$ distance matrix, shape $L \times L$ \\
$\boldsymbol{D}(\boldsymbol{X}), \boldsymbol{D}(\boldsymbol{\phi}, \boldsymbol{\psi})$ & Realization of $\boldsymbol{D}$ for given   coordinates or angles \\
$\pi(\boldsymbol{\phi}, \boldsymbol{\psi}; \boldsymbol{\Theta})$ & Prior over dihedral angles (von Mises) \\
$\pi(\boldsymbol{D}; \boldsymbol{\Gamma})$ & Prior over pairwise distances \\
$p(\boldsymbol{D}; \boldsymbol{\Lambda})$ & Target PDF of $\boldsymbol{D}$, used for PK update \\
\hline
Synthetic model (von Mises random walk, VRW) & \\
\hline
$N$ & Number of random walk steps\\ 
$\boldsymbol{\chi}$ & 2D coordinates of VRW starting at $(0,0)$, shape $(N+1) \times 2$ \\
$\boldsymbol{\theta}$ & Angles of a VRW, shape $N$ \\
$d$ & Origin-to-end distance of a VRW, $d \in \mathbb{R}^+ $ \\
$d(\boldsymbol{\chi}), d(\boldsymbol{\theta})$ & Realization of $d$ for given VRW 2D coordinates or angles \\
$\pi_{\textrm{VRW}}(\boldsymbol{\theta}; \mu, \kappa, N)$ & Prior over VRW angles (von Mises) \\
$\pi_{\textrm{VRW}}(d; \kappa, N)$, Stephens$(d; \kappa, N)$ & Prior over VRW origin-to-end distance $d$ \\
ScaledBeta$(d; \alpha, \beta, N)$ & Target PDF of $d$, used for PK update \\
\hline
PMF and PK summary & \\
\hline
$\boldsymbol{\omega}$ & A {\em fine-grained} random variable (e.g., $\boldsymbol{X}$) \\
$\boldsymbol{\xi}$ & A {\em coarse-grained} or {\em collective} random variable (e.g., $\boldsymbol{D}$) \\
$\boldsymbol{\xi}(\boldsymbol{\omega})$ & Realization of $\boldsymbol{\xi}$ given $\boldsymbol{\omega}$ (e.g., $\boldsymbol{D}(\boldsymbol{X})$) \\
J-condition & Jeffrey condition, posterior $p(\boldsymbol{\omega} \mid  \boldsymbol{\xi})=\pi(\boldsymbol{\omega} \mid  \boldsymbol{\xi})$ \\
\rule{0pt}{3ex}PMF; maps $p(\boldsymbol{\omega})$ to $p(\boldsymbol{\xi})$& $p(\boldsymbol{\xi})=\int \delta(\boldsymbol{\xi}-\boldsymbol{\xi}(\boldsymbol{\omega}))p(\boldsymbol{\omega})\textrm{d}\boldsymbol{\omega} $ \\[1.1ex]
PK; updates $\pi(\boldsymbol{\omega})$ with $p(\boldsymbol{\xi})$  & $p(\boldsymbol{\omega}) = \int \pi(\boldsymbol{\omega}\mid \boldsymbol{\xi}) p(\boldsymbol{\xi})\textrm{d}\boldsymbol{\xi}$ \\ [1.1ex]
Reference ratio formulation of PK & $p(\boldsymbol{\omega})=\int\frac{\delta(\boldsymbol{\xi}-\boldsymbol{\xi}(\boldsymbol{\omega}))\pi(\boldsymbol{\omega})}{\pi(\boldsymbol{\xi})}p(\boldsymbol{\xi})\textrm{d}\boldsymbol{\xi}=\frac{p(\boldsymbol{\xi}(\boldsymbol{\omega}))}{\pi(\boldsymbol{\xi}(\boldsymbol{\omega}))}\pi(\boldsymbol{\omega})$ \\[1.1ex] 
\hline
\end{tabular}
\end{table}

\FloatBarrier

\newpage

\section{AF1 AND THE SYNTHETIC MODEL}\label{sec:appendix_comparison}

Table \ref{table:comparision} summarizes the conceptual correspondences between AF1 and our synthetic von Mises random walk (VRW) model. Both models share the same probabilistic structure: a prior distribution over (fine-grained) angular variables induces Euclidean coordinates, which in turn determine a (coarse-grained) distance variable. The map from angles to distances is many-to-one in both cases. In AF1, dihedral angles generate 3D coordinates, $\boldsymbol{X}$, and corresponding $C\beta - C\beta$ distances; in the synthetic model, VRW angles generate 2D coordinates and an origin-to-end distance or {\em resultant length}, $d$.
This analogy ensures that the PK update plays the same role in both cases, i.e., an angular prior is updated with distance information, allowing us to study the probabilistic foundation of AF1 in a tractable and interpretable setting.

\begin{table}[H]
\centering
\caption{Comparison of AF1 and the synthetic model.}
\begin{tabular}{lll}
\hline
\textbf{AF1} & \textbf{Synthetic model} & \textbf{Correspondence} \\
\hline
Dihedral angles, ($\boldsymbol{\phi}, \boldsymbol{\psi}$) & VRW angles, $\boldsymbol{\theta}$ & Prior over angles (von Mises) \\
3D atomic coordinates, $\boldsymbol{X}$ & 2D VRW coordinates, $\boldsymbol{\chi}$ & Coordinates generated by angles \\
Pairwise distances, $\boldsymbol{D}$ & Resultant length, $d$ & PK update with distance information \\
\hline
\end{tabular}
\label{table:comparision}
\end{table}

\FloatBarrier

\section{WHITWORTH'S HORSES}\label{sec:appendix_whitworth}

This simple application of PK is discussed by Diaconis and Zabell  \citep{diaconis1982updating}. Three horses,
A, B, and C, have prior winning probabilities $\pi(A)=1/2$, $\pi(B)=1/4$,
and $\pi(C)=1/4$, respectively. New evidence increases the probability
of A winning from $\pi(A)=1/2$ to $p(A)=2/3$, without providing
any further information on the other horses. Standard Bayesian conditioning
cannot apply, as no specific event has occurred. Instead, PK prescribes
an update that satisfies the new marginal, $p(A)=2/3$ and $p(A^c)=1/3$, while preserving
the conditional probabilities, $\pi(B\mid A^c)=\pi(C\mid A^c)=1/2$. This
yields the updated posterior probabilities, 
\begin{align}
p(B) &=\pi(B\mid A^c)p(A^c)=1/2 \cdot 1/3=1/6, \\
p(C) &=\pi(C\mid A^c)p(A^c)=1/2 \cdot 1/3=1/6,
\end{align} 
while previously $\pi(B)=\pi(C)=1/4$.

\section{LOCAL NATURE OF AF1 PRIOR AND J-CONDITION}\label{sec:appendix_J}

\paragraph{Motivation} As the dihedral angles are sufficient to generate 3D structures, one might ask why it is necessary to add evidence on pairwise distances. It is well known that probabilistic models based on internal coordinates, such as dihedral angles, struggle to reproduce realistic distances between amino acids that are far apart in sequence position~\citep{arts2024internal}. This is due to the highly structured covariance induced by the chain geometry: small local perturbations in dihedral angles propagate in a correlated way through the backbone. 

\paragraph{Experiments} To illustrate this in the context of AF1, we sampled dihedral angles from the prior and reconstructed the corresponding 3D coordinates. AF1 code that can be executed for CASP13 targets is available from GitHub.\footnote{\scriptsize\url{https://github.com/google-deepmind/deepmind-research/tree/master/alphafold_casp13}} AF1 was applied to CASP13 target TS1019s2, corresponding to the PT-VENN domain-containing protein, Protein Data Bank Identifier 8EY4, chain B. For each amino acid, the prior is specified as probabilities over the Ramachandran plot, binned at $10^{\circ} \times 10^{\circ}$ resolution, obtained from a deep model (see Section \ref{sec:AF1's-Potential-Re-evaluated}) \cite{af1_nature_2020}. For each amino acid, we picked a random bin according to the specified probabilities, and set the dihedral angles to the bin's centroid. We then construct the 3D coordinates of the $N,C\alpha,C$ backbone from the sampled dihedral angles. The sampled structures were superimposed on the true structure using a standard algorithm \cite{kabsch1976, Dryden2016}. 
\begin{figure*}[tb]
\centering
\includegraphics[viewport=30.0156bp 0bp 360.156bp 317.167bp,clip,height=6cm]{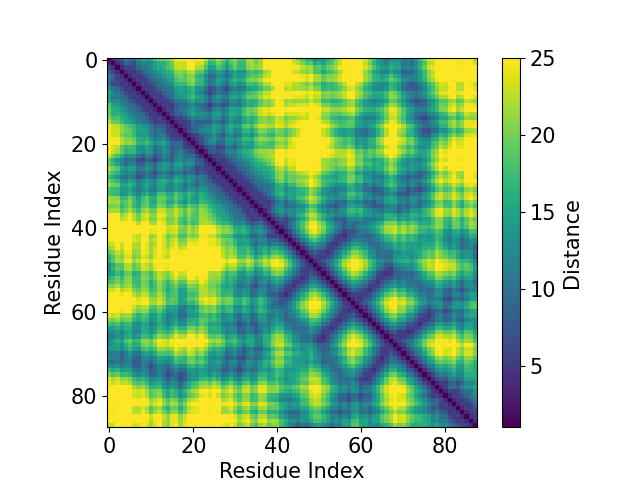}
\includegraphics[viewport=50.0234bp 0bp 417.781bp 317.167bp,clip,height=6cm]{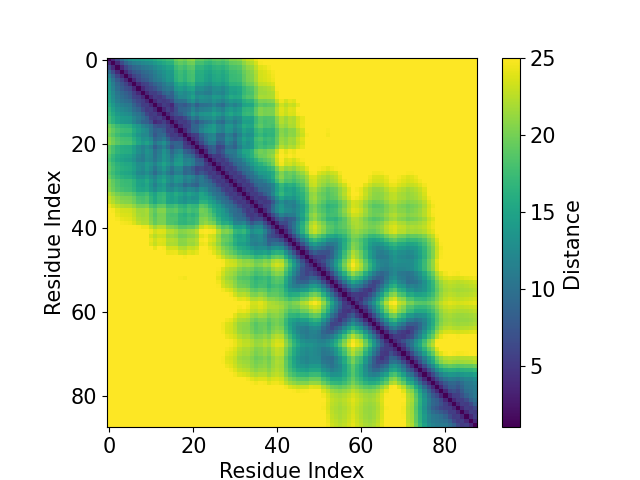}
\par
\caption{Distogram of the native structure (\textbf{left}) and the averaged distogram over 100 samples from the AF1 prior (\textbf{right}) for the PT-VENN domain-containing protein (Protein Data Bank Identifier 8EY4, Chain B). Distances are in Ångström. The prior accurately captures short-range distances between amino acids neighboring in sequence (i.e., distances close to the diagonal in the distogram on the right), but not long-range distances between amino acids far apart in the sequence, which necessitates updating the prior with distance information. \label{fig:distograms}}
\end{figure*}
\begin{figure*}
\centering
\includegraphics[height=6cm]{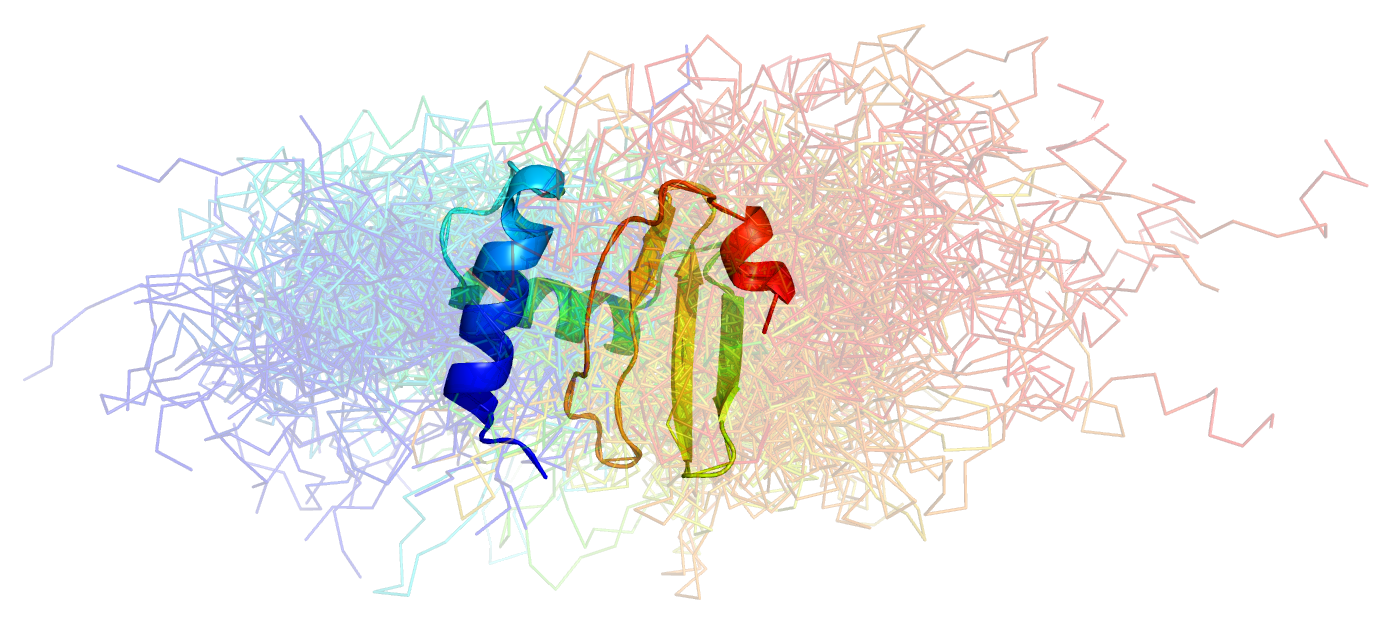}
\par
\caption{100 samples from the AF1 prior (transparent, thin ribbons)  superimposed on the cartoon representation of the native structure of the PT-VENN domain-containing protein (Protein Data Bank Identifier
8EY4, Chain B). In all cases, the N-terminus is shown in blue and C-terminus in red. Note how the samples from the prior are spatially extended compared to the compact native structure, which is representative for probabilistic models based on dihedral angles \citep{arts2024internal}. Figure made with PyMol \cite{PYMOL}.\label{fig:AF1_3D_samples}}
\end{figure*}
\paragraph{Results and discussion} Averaging over 100 sampled 3D structures, we visualized the resulting $C\alpha-C\alpha$ distogram and compared it to that of the native structure (see Figure~\ref{fig:distograms}). The prior captures distances between residues that are close together in the sequence well (i.e., close to the diagonal in Figure \ref{fig:distograms}), but fails to produce correct distances for the other cases, generating structures that are insufficiently compact. This is also visible in Figure ~\ref{fig:AF1_3D_samples}, where the ensemble of 3D structures sampled from the AF1 prior deviates significantly from the compact, folded conformation.
This result illustrates AF1's adopted J-condition, $p(\boldsymbol{\phi}, \boldsymbol{\psi} \mid \boldsymbol{D}) = \pi(\boldsymbol{\phi}, \boldsymbol{\psi} \mid \boldsymbol{D})$, and the corresponding PK update, which performs the minimal perturbation of the locally valid prior necessary to match the marginal distribution over the distances, $p(\boldsymbol{D})$. 

\section{PRACTICAL ASPECTS OF THE AF1 POTENTIAL\label{sec:appendix_approx}}

Our statistical framework explains  AF1's potential as the update of a prior over dihedral angles with information on pairwise distances. For practical reasons, the application of PK by AF1 is heuristic in a few respects, which we clarify here. 

First,
the reference distribution over the coarse grained variable, $\pi(\boldsymbol{D}(\boldsymbol{\phi},\boldsymbol{\psi});\boldsymbol{\Gamma})$,
is not directly obtained from the prior over the fine-grained variable,
$\pi(\boldsymbol{\phi},\boldsymbol{\psi};\boldsymbol{\Theta})$. Rather,
it is approximated by training a model that uses as input only the
sequence length, $L$, and the positions of the Glycine
residues, $\mathbb{I}_{\textrm{Gly}}(\boldsymbol{s})$. 

Second, AF1 does not use proper PDFs over the distance matrix, $\boldsymbol{D}$.
For $n$ points in $\mathbb{R}^{d}$, the $n(n-1)/2$ pairwise distances
cannot vary independently: they must satisfy geometric constraints
to correspond to a valid 3D configuration. As discussed by \cite{clifford1985distances}, valid distances lie on a
smaller manifold of dimension $nd-d(d+1)/2$. By modeling probability
distribution over $\boldsymbol{D}$ as a product of independent marginals, AF1 defines an approximate distribution that
is \emph{singular}, i.e., it may place nonzero density on geometrically
invalid configurations. 

 These choices reflect pragmatic compromises that ultimately preserve the core principle of PK: updating a prior distribution concerning a fine-grained variable using soft evidence on a coarse-grained variable. In addition, minimization of the AF1 potential as a function of the dihedral angles 
 ensures realizable 3D coordinates and a corresponding valid distance matrix.

\section{PROTEIN EXPERIMENT}\label{sec:3D_experiment}

\begin{figure*}
\centering
\includegraphics[width=0.59\textwidth]{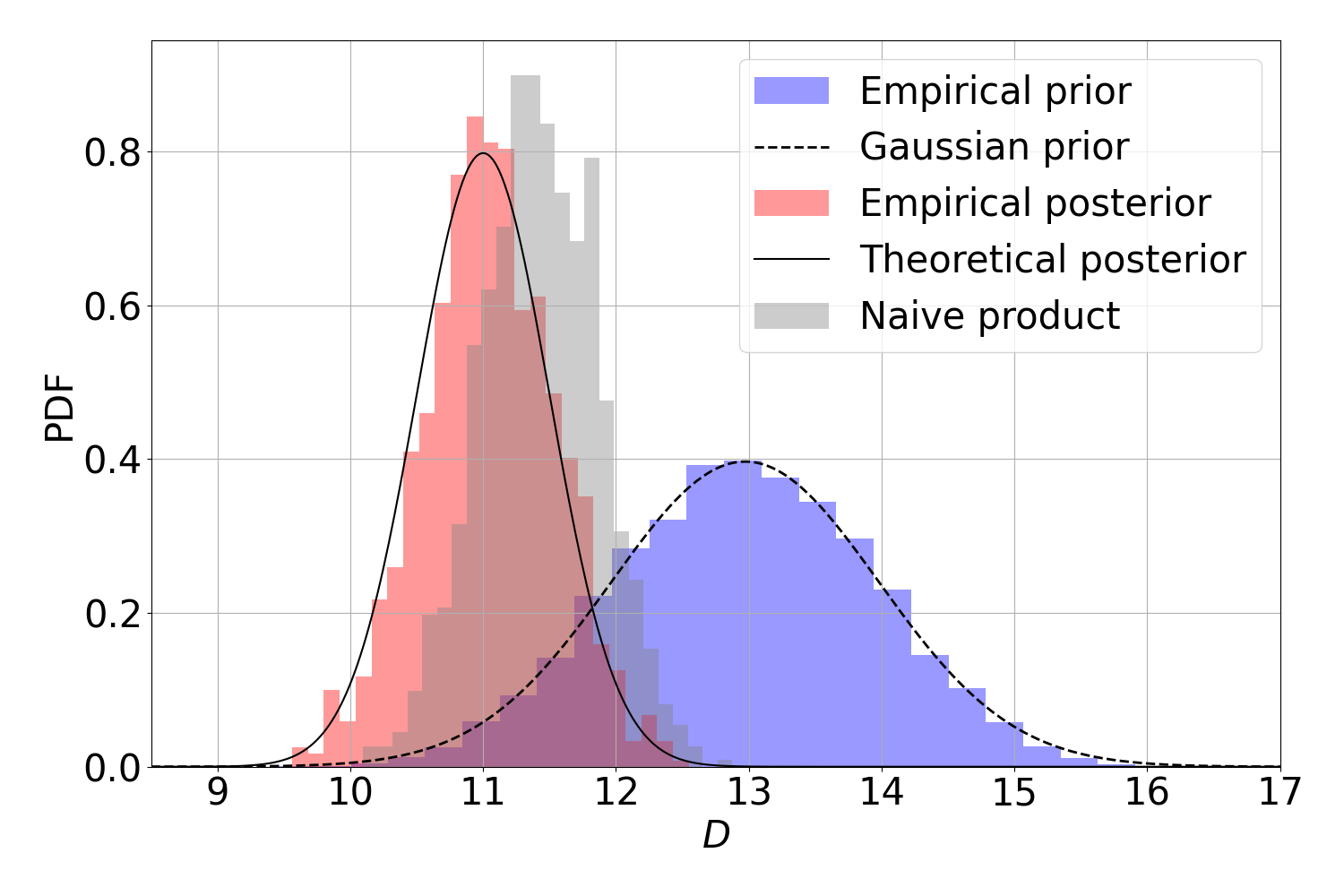}
\hfill
\includegraphics[width=0.39\textwidth]{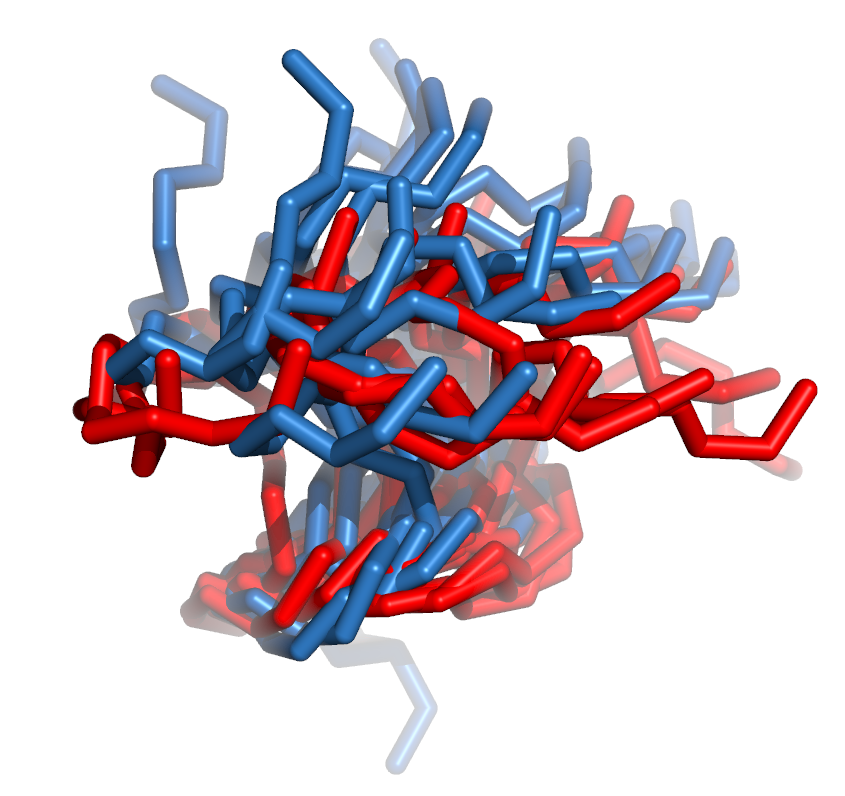}
\par
\caption{PK update applied to a 3D protein model (von Mises angular prior, eight amino acids). \textbf{(left)}  Empirical PDFs of the $C\alpha-C\alpha$ start–to–end distance,
$D$, for the unmodified angular prior (blue histogram), the PK–updated posterior (red histogram), and a naive product of
the Gaussian target with the empirical prior, i.e., without dividing by the reference distribution (gray histogram).
Also shown are the Gaussian distribution that approximates $\pi(D)$ (dashed black), estimated from prior samples, and the
theoretical posterior
(i.e., the desired target Gaussian distribution, $p(D)$; solid black), showing good agreement with the actual PK posterior. 
\textbf{(right)} 3D conformations from the prior (blue) and the PK–updated distribution (red), generated by the PNERF algorithm from the dihedral angles. Figure made with PyMol \citep{PYMOL}.}
\label{fig:helix}
\end{figure*}

\paragraph{Motivation}

To demonstrate that PK readily extends to realistic
three–dimensional systems of relevance to protein structure prediction,
we applied the framework to a model that represents a small protein consisting of eight amino acids.
The experimental design closely parallels the two-dimensional von Mises random walk (VRW) of Section~\ref{sec:experiments},
but now uses dihedral angles describing the 3D coordinates of the protein backbone.
As in AF1, a generative model that generates 3D Euclidean coordinates from (von Mises) angular priors
is updated with evidence on inter-atomic distances. 
Recall that the 2D VRW synthetic model was chosen because the distance prior $\pi(d)$ associated with the angular prior is analytically available (see Section \ref{sec:experiments}, Setup). This is not the case for the protein model. Thus, the distance prior must be estimated empirically from samples generated from the angular prior. Remarkably, a simple Gaussian fit to these samples provides a sufficient approximation to recover the correct posterior, highlighting the robustness of the PK formulation. In the next paragraph, we describe the protein model in detail. 
 
\paragraph{Setup}

As before, let the $L \times 4 \times 3$ tensor $\boldsymbol{X}$ contain the atomic coordinates ($N,C\alpha,C,C\beta$) of a protein backbone, and let $\boldsymbol{X}(\boldsymbol{\phi}, \boldsymbol{\psi})$ be the realization of $\boldsymbol{X}$ at $(\boldsymbol{\phi}, \boldsymbol{\psi})$. In this section, we will consider a protein consisting of eight amino acids (i.e., $L=8$). Following AF1, bond lengths and bond angles are fixed to ideal values, and all peptide bonds were set to the \emph{trans} conformation, without any loss of generality. Again following AF1, the prior over each dihedral angle follows a von Mises distribution. For our protein model, the prior over the dihedral angles of the $i$-th amino acid is
\begin{equation}\label{eq:protein_prior}
\phi_i\sim\mathrm{vM}(\mu_\phi,\kappa_\phi),\qquad
\psi_i\sim\mathrm{vM}(\mu_\psi,\kappa_\psi),
\end{equation}
where $\mu_\phi=-60^{\circ}$, $\mu_\psi=-40^{\circ}$, which  
corresponds to a canonical right-handed $\alpha$-helix, and $\kappa_\phi=\kappa_\psi=20$. This prior distribution corresponds to a probability distribution over the space of $\alpha$-helix-like 3D conformations. As is the case for AF1, we will update this angular prior with pairwise distance information using PK. 
Concretely, we use PK to alter the PDF of the distance\footnote{We use a single distance ($D$) instead of all pairwise distances ($\boldsymbol{D}$) without any loss of generality because (a) this makes analysis easier and (b) as formulating non-singular PDFs over distance matrices is somewhat non-trivial \citep{clifford1985distances}, we leave this for future work. We use a capital letter despite the scalar nature of $D$ to emphasize its conceptual correspondence to the distance matrix $\boldsymbol{D}$.} $D$ between the first and the last $C\alpha$ atoms, 
\begin{equation}\label{eq:D_peptide}
D(\boldsymbol{\phi}, \boldsymbol{\psi})=\lVert\mathbf{X}(\boldsymbol{\phi}, \boldsymbol{\psi})_{0,a}-\mathbf{X}(\boldsymbol{\phi}, \boldsymbol{\psi})_{-1,a}\rVert,
\end{equation}
where $D(\boldsymbol{\phi}, \boldsymbol{\psi})$ is the realization of $D$ at $(\boldsymbol{\phi}, \boldsymbol{\psi})$ and $a$ is the index of the $C\alpha$ atom. In contrast to the tractable synthetic model based on the VRW, the induced prior distribution over the distance, $\pi(D)$, is not available for our protein model. Therefore, $\pi(D)$
was approximated by a Gaussian,
\begin{equation}\label{eq:protein_prior_D}
\pi(D)\approx\mathcal{N}(\mu_\pi,\sigma_\pi^2),\qquad
\mu_\pi=12.97~\text{\AA},\;
\sigma^2_\pi=1.02~\text{\AA},
\end{equation}
where $(\mu_\pi,\sigma_\pi)$ were robustly estimated (using the median and median absolute deviation, respectively) from $10^4$ samples from the prior. Figure~\ref{fig:helix} (left) shows the histogram of the samples from the prior (blue) and the fitted Gaussian (dashed black line). 
As target distribution, we chose a Gaussian distribution that imposes a smaller mean distance and a sharper distribution,
\begin{equation}\label{eq:protein_target}
p(D)=\mathcal{N}(\mu_D,\sigma_D^2),
\qquad
\mu_D=11~\text{\AA},\;
\sigma^2_D=0.25~\text{\AA}.
\end{equation}
As is the case for AF1 and the VRW, the conditional distribution, i.e.,  $p(\boldsymbol{\phi}, \boldsymbol{\psi} \mid D)$, is not available. Therefore, we again use the reference formulation of PK, which does not require the conditional, but uses a ratio of marginals. Following Eq.~(\ref{eq:rrm}), the PK update is given by
\begin{align}
p(\boldsymbol{\phi}, \boldsymbol{\psi})
&=
\frac{p(D(\boldsymbol{\phi}, \boldsymbol{\psi}))}{\pi(D(\boldsymbol{\phi}, \boldsymbol{\psi}))}\,
\pi(\boldsymbol{\phi}, \boldsymbol{\psi}),
\label{eq:helix_pk}
\end{align}
which after filling in the concrete PDFs (\ref{eq:protein_prior}, \ref{eq:protein_prior_D}, \ref{eq:protein_target}) results in
\begin{align}
p(\boldsymbol{\phi}, \boldsymbol{\psi})
&=
\frac{\mathcal{N}(D(\boldsymbol{\phi}, \boldsymbol{\psi}) \mid \mu_D, \sigma_D)}{\mathcal{N}(D(\boldsymbol{\phi}, \boldsymbol{\psi}) \mid \mu_\pi, \sigma_\pi)}\,
\prod_{i=0}^{L-1}\mathrm{vM}(\phi_i \mid \mu_\phi,\kappa_\phi)
\mathrm{vM}(\psi_i \mid \mu_\psi,\kappa_\psi).
\label{eq:helix_pk_concrete}
\end{align}

Similarly to the VRW model, we sample the model using the No-U-Turn Sampler (NUTS) \citep{hoffman2014no} as implemented in Pyro~\citep{bingham2019pyro}.  We use
1000 warm-up and 1000 posterior samples, and set the maximum tree depth to 6. 3D coordinates were calculated from the dihedral angles using the Parallelized Natural Extension Reference Frame (PNERF) algorithm \citep{parsons2005practical,alquraishi2019parallelized},\footnote{We used the PyTorch implementation available from \url{https://github.com/FelixOpolka/pnerf-pytorch}.} i.e.,  $\boldsymbol{X}(\boldsymbol{\phi}, \boldsymbol{\psi})$ in (\ref{eq:D_peptide}) is $\textrm{PNERF}(\boldsymbol{\phi}, \boldsymbol{\psi})$. Execution takes under two minutes on a standard laptop with an Intel i7 CPU.

\paragraph{Results and evaluation}

Figure~\ref{fig:helix} (left) compares the empirical distributions of $D$
obtained under the prior (blue) and the posterior obtained from PK (red).
The PK posterior visually coincides well with the imposed target distribution (Gaussian given by \ref{eq:protein_target}, black line). As in the VRW case, we further evaluate the fit between the posterior samples and the target distribution using the KS test, including every fifth NUTS sample to reduce autocorrelation. Across ten NUTS runs, the KS statistics range from 0.03 to 0.10, with a median p-value equal to 0.41, confirming a good fit.   

\paragraph{Ablation experiment}

In contrast and as expected, omitting the reference distribution and using a naive product, i.e., $p(D(\boldsymbol{\phi}, \boldsymbol{\psi}))
\pi(\boldsymbol{\phi}, \boldsymbol{\psi})$, produced a markedly poorer fit (gray histogram in Figure~\ref{fig:helix}, left). The KS statistics ranged from $0.33$ to $0.40$ with $p<10^{-18}$ in all cases, confirming the poor fit. As in the case of the VRW, the ablation study confirms the importance of the reference distribution. 

\paragraph{Discussion}

This experiment demonstrates and confirms that the PK update is applicable to a protein model, despite the need for an estimation of the reference distribution: even though the prior marginal $\pi(D)$ is unavailable in closed form, a simple Gaussian approximation estimated from prior samples suffices to reproduce a matching posterior distribution.


\end{document}